\documentclass[manuscript, screen]{acmart}
\AtBeginDocument{%
  }

\setcopyright{acmlicensed}
\copyrightyear{2018}
\acmYear{2018}
\acmDOI{XXXXXXX.XXXXXXX}
\acmConference[Conference acronym 'XX]{Make sure to enter the correct
  conference title from your rights confirmation email}{June 03--05,
  2018}{Woodstock, NY}
\acmISBN{978-1-4503-XXXX-X/2018/06}

\usepackage{makecell}
\usepackage{pifont}
\usepackage{booktabs}
\usepackage{subcaption}
\usepackage{multirow}
\usepackage{xurl}

\usepackage{tikz}
\usetikzlibrary{shapes, positioning, patterns, mindmap, shadows, shapes.geometric}

\definecolor{red}{rgb}{0.957,0.498,0.447}
\definecolor{lightred}{rgb}{0.965,0.702,0.675}
\definecolor{green}{rgb}{0.553,0.824,0.773}
\definecolor{lightgreen}{rgb}{0.729,0.890,0.863}
\definecolor{blue}{rgb}{0.498,0.698,0.835}
\definecolor{lightblue}{rgb}{0.702,0.820,0.906}
\definecolor{orange}{rgb}{0.969,0.718,0.427}
\definecolor{lightorange}{rgb}{0.988,0.831,0.631}
\definecolor{purple}{rgb}{0.749,0.737,0.855}
\begin{document}

\title{A Survey on Unlearning in Large Language Models}

\author{Ruichen Qiu}
\email{qiuruichen20@mails.ucas.ac.cn}
\affiliation{%
  \institution{School of Advanced Interdisciplinary Sciences, UCAS}
  \city{Beijing}
  \country{China}
}
\affiliation{%
  %\institution{Academy of Mathematics and Systems Science, Chinese Academy of Sciences}
  \institution{Academy of Mathematics and Systems Science, CAS}
  \city{Beijing}
  \country{China}}

\author{Jiajun Tan}
\affiliation{%
  %\institution{Institute of Computing Technology, Chinese Academy of Sciences}
  \institution{Institute of Computing Technology, CAS}
  \city{Beijing}
  \country{China}}

\author{Jiayue Pu}
\affiliation{%
  \institution{University of Chinese Academy of Sciences}
  \city{Beijing}
  \country{China}}

\author{Honglin Wang}
\affiliation{%
  %\institution{Institute of Computing Technology, Chinese Academy of Sciences}
  \institution{Institute of Computing Technology, CAS}
  \city{Beijing}
  \country{China}}

\author{Xiao-Shan Gao}
\affiliation{%
  %\institution{Academy of Mathematics and Systems Science, Chinese Academy of Sciences}
  \institution{Academy of Mathematics and Systems Science, CAS}
  \city{Beijing}
  \country{China}}
\email{xgao@mmrc.iss.ac.cn}

\author{Fei Sun}
\affiliation{%
  %\institution{Institute of Computing Technology, Chinese Academy of Sciences}
  \institution{Institute of Computing Technology, CAS}
  \city{Beijing}
  \country{China}}
\email{sunfei@ict.ac.cn}

\renewcommand{\shortauthors}{Qiu et al.}

\begin{abstract}
    % 100 words version
    % Large Language Models (LLMs) demonstrate remarkable capabilities, but their training on massive corpora poses risks from memorized sensitive information. 
    % To mitigate these risks, unlearning becomes an essential paradigm for the selective removal of specific knowledge.
    % This survey provides a systematic review of over 180 papers on LLM unlearning published since 2021.
    % It introduces a novel taxonomy of unlearning methods based on the training stage of intervention.
    % For evaluations, it provides multidimensional comparisons of datasets and metrics.
    % By discussing current challenges and future directions, this survey aims to advance the field of LLM unlearning and the development of secure AI systems.
    %
    %
    % normal version
    Large Language Models (LLMs) demonstrate remarkable capabilities, but their training on massive corpora poses significant risks from memorized sensitive information. 
    To mitigate these issues and align with legal standards, unlearning has emerged as a critical technique to selectively erase specific knowledge from LLMs without compromising their overall performance.
    This survey provides a systematic review of over 180 papers on LLM unlearning published since 2021.
    First, it introduces a novel taxonomy that categorizes unlearning methods based on the phase in the LLM pipeline of the intervention.
    This framework further distinguishes between parameter modification and parameter selection strategies, thus enabling deeper insights and more informed comparative analysis. 
    Second, it offers a multidimensional analysis of evaluation paradigms.
    For datasets, we compare 18 existing benchmarks from the perspectives of task format, content, and experimental paradigms to offer actionable guidance. 
    For metrics, we move beyond mere enumeration by dividing knowledge memorization metrics into 10 categories to analyze their advantages and applicability, while also reviewing metrics for model utility, robustness, and efficiency.
    By discussing current challenges and future directions, this survey aims to advance the field of LLM unlearning and the development of secure AI systems.
\end{abstract}

\ccsdesc[500]{Information systems~Language models}
\ccsdesc[500]{Security and privacy}

%\keywords{Do, Not, Use, This, Code, Put, the, Correct, Terms, for,
%  Your, Paper}
\keywords{Machine Unlearning, Large Language Models}

\received{20 February 2007}
\received[revised]{12 March 2009}
\received[accepted]{5 June 2009}

\maketitle

% !TeX root = ../main_csur.tex
\section{Introduction}
Large Language Models (LLMs) have significantly transformed research paradigms in natural language processing while enabling a diverse array of practical applications. 
These capabilities arise from training on extensive textual corpora, which allows the models to encode substantial knowledge within their parameters. 
However, this capacity also introduces critical risks. 
For instance, personally identifiable information memorized during training can be extracted through privacy attacks, raising concerns under data protection regulations such as ``right to be forgotten'' \citep{voigt2017eu,regulation2016regulation}. 
Similarly, unauthorized use of copyright materials in training data can expose model providers to legal challenges~\citep{Wei2024Evaluating}. 
Moreover, LLMs can internalize knowledge that facilitates malicious activities~\citep{Li2024WMDP,lang2025beyond}, and jailbreak attacks can elicit the generation of harmful or illegal content. 
In light of these concerns, selectively erasing specific knowledge from LLMs has emerged as a necessary step toward enhancing their security, reliability, and regulatory compliance.

%While retraining models from scratch could theoretically address these issues, it is computationally expensive and impractical for large-scale models.
One potential solution is to retrain LLMs from scratch after removing problematic data. However, this approach is computationally expensive and impractical for large-scale models. \textbf{Machine unlearning} \citep{Cao2015Making} offers a more efficient alternative, which aims to develop algorithms to selectively remove the influence of specific training data while preserving the overall performance of the model on retained data.
%\textbf{Machine Unlearning}~\citep{Cao2015Making}, which aims to develop algorithms that efficiently remove the influence of particular training data without degrading performance on retained data, offers a foundational and versatile approach.
In the context of LLMs, the distinctive autoregressive next-token prediction mechanism~\citep{yao2024machine} has motivated extensive research into unlearning methods specifically designed for these models. 
This survey narrows its focus to address unlearning techniques tailored for large-scale generative language models, which are predominantly used for generative tasks rather than classification.\footnote{Some LLM unlearning works also considered classification tasks in natural language processing~\citep{Pawelczyk2024Context,Bhaila2024Soft,Chen2023Unlearn}, but they are not the focus of this survey.}

Several existing surveys touch upon LLM unlearning.
Most of them either adopt a broader scope~\citep{cooper2024machine,zhou2024limitations,liu2024machine,xu2024machine}, concentrate on specialized aspects~\citep{barez2025open,zhang2024right,Si2023Knowledge,qu2024frontier}, or lack coverage of extensive research published after October 2024~\citep{blanco2025digital,le-khac2025survey}.
To address these gaps, we provide a comprehensive and up-to-date overview by systematically reviewing more than 180 papers published since 2021.\footnote{Some articles were retrieved from public repositories such as \url{https://github.com/chrisliu298/awesome-llm-unlearning}.}
In contrast to other recent surveys on LLM unlearning~\citep{ren2025sok,geng2025comprehensive}, our work introduces a novel taxonomy and delivers an in-depth analysis, with specific contributions outlined below.

\textbf{(1) A novel taxonomy of unlearning methods (Section~\ref{sec:methods}).}
%We categorize unlearning approaches based on the training stage at which they are applied: training time, post-training, and inference time. 
We categorize unlearning approaches based on their execution timing: training time, post-training, and inference time.
Compared to alternative taxonomies based on unlearning objectives or intentions, our framework provides a clearer organizational structure with two advantages.
First, it distinguishes between parameter modification and parameter selection strategies, offering a flexible conceptual space for their integration, and thereby enabling deeper insights.
For example, while RMU~\citep{Li2024WMDP} is grouped under categories such as ``localized parameter modification'' in the prior survey, our taxonomy separately analyzes its loss design in the SFT section (\ref{sec:ft_sft}) and its parameter selection mechanism in the parameter localization section (\ref{sec:ft_local}).
Second, our taxonomy is usage-oriented, enabling the informed selection and comparative analysis of unlearning methods according to specific operational scenarios.

%Second, as a usage oriented classification, our taxonomy facilitates the selection and comparison of unlearning methods based on specific usage scenarios.
% First, we separate the methods of parameter modification and parameter selection, providing flexible space for the combination of the two to provide readers with deeper insights.
% For example, RMU~\citep{Li2024WMDP} is classified in ``local weight modification'' or similar sections in~\citep{geng2025comprehensive,blanco2025digital}, but we classify its loss design and parameter selection methods in the SFT section (\ref{sec:ft_sft}) and parameter localizing section (\ref{sec:ft_local}), respectively.
% Second, our taxonomy helps select and compare unlearning methods according to the required usage scenarios.
%, since some full-parameter methods can also be applied to part of the parameters or by incorporating LoRA adapters to apply to extra parameters.

\textbf{(2) Multidimensional analysis of evaluations (Section~\ref{sec:eval}).}
Instead of merely enumerating existing datasets and metrics, we provide a multidimensional analysis for both datasets and metrics.
For datasets, through a comparison from the perspectives of task format, content, and experimental paradigms, we evaluate the characteristics of 18 existing benchmarks, offering actionable guidance for researchers.
For metrics, from the goal of LLM unlearning, we divide knowledge memorization metrics into 10 categories to analyze their advantages and applicability, along with commonly used metrics for model utility, robustness, and efficiency.

\textbf{(3) Discussion of Challenges and Future Directions (Section~\ref{sec:dis}).}
We provide an in-depth discussion of current challenges in LLM unlearning, including the lack of a strict and consistent definition, the variation of impacts across languages and data, and the difficulties of real-world implementation.
%We provide an in-depth discussion of current challenges in LLM unlearning from three aspects, including the lack of consistent and strict definition, impact varying across different languages and data, and implement in real-world scenarios.
Furthermore, we outline prospective research directions, aiming to advance the field of LLM unlearning and contribute to more responsible AI systems.

\section{Backgrounds}

\subsection{Machine Unlearning in LLMs}
\label{sec:def}
\begin{figure}[t]
    \centering
    \footnotesize
    \begin{tikzpicture}
        %\node at (0,0.5) [rectangle, dotted, draw=black, fill=none, minimum width=380pt, minimum height=70pt, rounded corners=10pt, line width=1.5pt] {};
        \node at (-5,4) {\includegraphics[width=1cm]{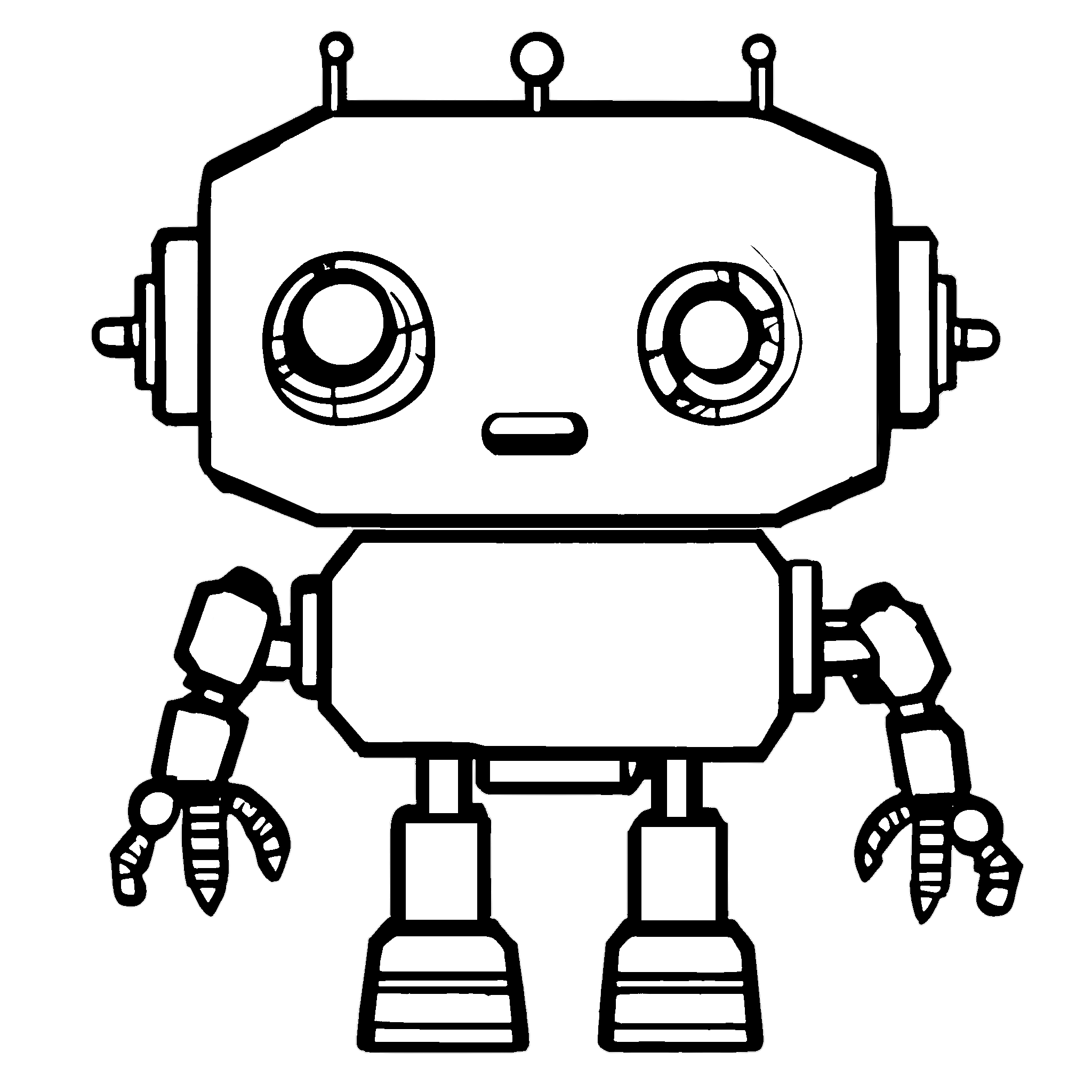}};
        \node (llm_target) at (-5,3) [rectangle, draw=black, fill=none, minimum width=2.6cm, minimum height=0.5cm, inner sep=0pt, rounded corners=5pt, line width=1pt] {};
        \node at (-5,2.5) [] {\footnotesize Original Model $\mathcal{M}$};
        \node at (-6,3) [regular polygon, regular polygon sides=4, draw=black, fill=none, minimum width=.4cm, inner sep=0pt] {};
        \node at (-5.5,3) [regular polygon, regular polygon sides=4, draw=black, fill=none, minimum width=.4cm, inner sep=0pt] {};
        \node at (-5,3) [regular polygon, regular polygon sides=4, draw=black, fill=none, minimum width=.4cm, inner sep=0pt] {};
        \node at (-4.5,3) [regular polygon, regular polygon sides=4, draw=black, fill=none, minimum width=.4cm, inner sep=0pt] {};
        \node at (-4,3) [regular polygon, regular polygon sides=4, draw=black, fill=none, pattern=north east lines, minimum width=.4cm, inner sep=0pt] {};

        \node at (0,4) {\includegraphics[width=1cm]{figs/Survey.png}};
        \node (llm_final) at (0,3) [rectangle, draw=black, fill=none, minimum width=2.6cm, minimum height=0.5cm, inner sep=0pt, rounded corners=5pt, line width=1pt] {\small \textbf{?}};
        \node at (0,2.5) [] {\footnotesize Unlearned Model $\mathcal{M}_u$};
        \draw [thick,->,line width=1pt] ([xshift=2pt]llm_target.east) -- node(unlearning)[above,draw=none,fill=none] {\makecell{Unlearning\\algorithm $\mathcal{U}$}} ([xshift=-2pt]llm_final.west);
        
        \node at (5,4) {\includegraphics[width=1cm]{figs/Survey.png}};
        \node (llm_best) at (5,3) [rectangle, draw=black, fill=none, minimum width=2.6cm, minimum height=0.5cm, inner sep=0pt, rounded corners=5pt, line width=1pt] {};
        \node at (5,2.5) [] {\footnotesize Retrained Model $\mathcal{M}_r$};
        \node at (4,3) [regular polygon, regular polygon sides=4, draw=black, fill=none, minimum width=.4cm, inner sep=0pt] {};
        \node at (4.5,3) [regular polygon, regular polygon sides=4, draw=black, fill=none, minimum width=.4cm, inner sep=0pt] {};
        \node at (5,3) [regular polygon, regular polygon sides=4, draw=black, fill=none, minimum width=.4cm, inner sep=0pt] {};
        \node at (5.5,3) [regular polygon, regular polygon sides=4, draw=black, fill=none, minimum width=.4cm, inner sep=0pt] {};
        \draw [thick,<->,line width=1pt] ([xshift=2pt]llm_final.east) -- node(vs)[above,draw=none,fill=none] {\makecell{Compare}} ([xshift=-2pt]llm_best.west);
    \end{tikzpicture}
    \vspace{-3pt}
    \caption{Illustration of an unlearning process. The box below the model represents the composition of the corresponding training set. The unlearn set $\mathcal{D}_u$ is represented by the shadow square and the retain set $\mathcal{D}_r$ is represented by the white square. An unlearning algorithm is applying on the initial target model to obtain the unlearned model $\mathcal{M}_u$. And the unlearned model is expected to approximate the retrained model $\mathcal{M}_r$.}
    \label{fig:unlearn_def}
\end{figure}

Within the standard framework of machine unlearning, we consider a dataset $\mathcal{D}$ and an \textit{original model} $\mathcal{M}$, parameterized by $\theta$, trained on $\mathcal{D}$.  
The subset of training data targeted for removal is denoted as the \textit{unlearn set} $\mathcal{D}_u \subset \mathcal{D}$, while the remainder constitutes the \textit{retain set} $\mathcal{D}_r = \mathcal{D} \setminus \mathcal{D}_u$.  
The objective of machine unlearning is to design an algorithm $\mathcal{U}$ that takes the original model $\mathcal{M}$ and the relevant data as input and produces an \textit{unlearned model} $\mathcal{M}_u$.  
This model $\mathcal{M}_u$ is intended to approximate the behavior of a \textit{retrained model} $\mathcal{M}_r$, which is trained exclusively on the retain set $\mathcal{D}_r$.
%\footnote{For notational convenience, we may use the model notations ($\mathcal{M}, \mathcal{M}_u, \mathcal{M}_r$) and their corresponding parameter notations ($\theta, \theta_u, \theta_r$) interchangeably in subsequent sections.}  
An illustration of the unlearning process is provided in Figure~\ref{fig:unlearn_def}.
In the context of LLM unlearning, we provide specific explanations from two aspects: (1) different types of unlearning request and (2) the goal of unlearning.

%These inherent characteristics cause LLM unlearning more complex, both in terms of the type of request and the goal.

\subsubsection{Type of Unlearning Request}
\label{sec:request}
\begin{figure}[t]
    \centering
    \vspace{-5pt}
    \begin{tikzpicture}
        \node (sample) at (-3.6,0.05) [rectangle, draw=blue, fill=none, minimum width=7.5cm, minimum height=3.7cm, inner sep=5pt, rounded corners=5pt, line width=1pt] {};
        \node at (-3.6,1.9) [fill=white,inner sep=2pt] {\textcolor{blue}{\Large \em Sample-level}};
        \node (privacy) at (-5.05,0.85) [rectangle, draw=blue, fill=lightblue!50, inner sep=5pt, rounded corners=5pt, line width=1pt, text width=3.9cm, align=left] {
            {\large \textbf{Privacy}}\\
            \footnotesize Anallise Ivory was born on November 8, 1990, and her Social Security Number is 900-55-1236.~\citep{ramakrishna2025lume}\\
            ~\vspace{-\baselineskip}
        };
        \node (copyright) at (-5.05,-0.85) [rectangle, draw=green, fill=lightgreen!50, inner sep=5pt, rounded corners=5pt, line width=1pt, text width=3.9cm, align=left] {
            \large\textbf{Copyright}\\
            \footnotesize ``There's more in the frying pan,'' said Aunt Petunia, turning eyes on her massive son.~\citep{Shi2024MUSE}\\
            ~\vspace{-\baselineskip}
        };
        \node (safety) at (-1.4,0) [rectangle, draw=purple, fill=purple!20, inner sep=5pt, rounded corners=5pt, line width=1pt, text width=2.4cm, align=left] {
            \large\textbf{Safety}\\
            \footnotesize This directory contains analyses for the FirmAE system.\textbackslash n\textbackslash n* \verb|`|fuzzer.py\verb|`|: This is a main script for testing command injection and buffer overflow vulnerability.~\citep{Li2024WMDP}\\
            ~\vspace{-0.05\baselineskip}
        };
        
        \node (entity) at (3.6,0.05) [rectangle, draw=red, fill=none, minimum width=6.5cm, minimum height=3.7cm, inner sep=0pt, rounded corners=5pt, line width=1pt] {};
        \node at (3.6,1.9) [fill=white,inner sep=2pt] {\textcolor{red}{\Large \em Entity-level}};
        \node (concrete) at (3.6,0.85) [rectangle, draw=red, fill=lightred!50, inner sep=5pt, rounded corners=5pt, line width=1pt, text width=5.9cm, align=left] {
            {\large \textbf{Concrete}}\\
            \footnotesize
            \textbf{Entity:} Stephen King\\
            \textbf{Samples:} Stephen King is a world-renowned American author of horror, suspense, supernatural fiction, ...~\citep{Jin2024RWKU}\\
            ~\vspace{-\baselineskip}
        };
        \node (abstract) at (3.6,-0.85) [rectangle, draw=orange, fill=lightorange!50, inner sep=5pt, rounded corners=5pt, line width=1pt, text width=5.9cm, align=left] {
            {\large \textbf{Abstract}}\\
            \footnotesize 
            \textbf{Entity:} Brute Force\\
            \textbf{Samples:} Adversaries may use brute force techniques to gain access to ...~\citep{lang2025beyond}\\
            ~\vspace{-\baselineskip}
        };
    \end{tikzpicture}
    \vspace{-3pt}
    \caption{Examples of different requests. We extract some fragments from the unlearn set of the corresponding work. At an entity level, in addition to the entity for unlearning, we also show generated samples of these entity, giving an illustration of converting entity-level unlearning to sample-level unlearning.}
    \label{fig:req_example}
\end{figure}
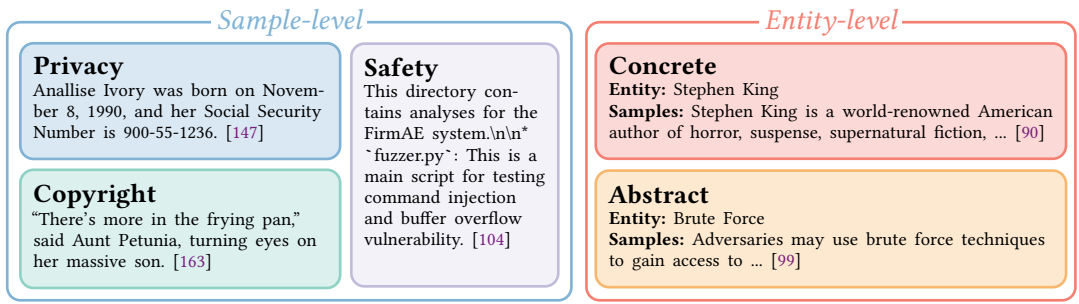
The predominant form of unlearning request operates at the sample level, requiring models to forget specific \textbf{text sequences} that contain sensitive information, thereby mitigating privacy~\citep{Tian2024Forget,ramakrishna2025lume}, copyright~\citep{Eldan2023Whos,Shi2024MUSE}, or safety risks~\citep{Li2024WMDP}. 
Examples of different samples are shown in Figure~\ref{fig:req_example}.
These sequences may consist of free-form text or structured question–answer pairs, as outlined in Table~\ref{tab:D&B}.

Beyond isolated samples, a growing number of work addresses entity-level unlearning, which aims at removing \textbf{all knowledge associated with a particular entity}. 
Entities may be concrete (e.g., individuals, books)~\citep{Jin2024RWKU,choi2025opt-out} or abstract (e.g., biases, capabilities)~\citep{lang2025beyond}, as depicted in Figure~\ref{fig:req_example}. 
Usually, this task is reduced to sample-level unlearning by constructing a corresponding unlearn set with samples related to the target entity. 
Compared to sample-level unlearning, it requires not only erasing memorized content but also managing inter-entity correlations, rendering it significantly more challenging.

% The main type of unlearning request is at the sample level, where the model need to unlearn specific \textbf{text sequences} that may contain sensitive information, posing privacy~\citep{Tian2024Forget,ramakrishna2025lume}, copyright~\citep{Eldan2023Whos,Shi2024MUSE} or safety risks~\citep{Li2024WMDP}. 
% Examples of different samples are shown in Figure~\ref{fig:req_example}.
% These sequences may be free-form or structured as question–answer pairs, as discussed in Section~\ref{sec:eval_task}.

% Moving beyond unlearning isolated samples, several works consider a new setting of entity level learning, which involves removing \textbf{all information related to a particular entity}.
% The entity can be concrete (e.g., individuals, books)~\citep{Jin2024RWKU,choi2025opt-out} or abstract (e.g., biases, capabilities)~\citep{lang2025beyond}, shown in Figure~\ref{fig:req_example}.
% This form of unlearning is usually converted into sample level unlearning by construct a unlearn set with lots of samples related to the target entity.
% However, it necessitates not only erasing memorized content but also appropriately handling inter-entity correlations, making it considerably more challenging than sample level unlearning.

%Given the extensive input and output spaces of LLMs, along with their diverse application scenarios, LLM unlearning requests can be broadly classified into sample level and entity level based on the form of the unlearn set:

%\textbf{(2) Text-level unlearning.}  

%\textbf{(3) Entity-level unlearning.}  

\subsubsection{Goals of Unlearning}
\label{sec:def_goal}
In traditional machine unlearning, the principal objective of the unlearned model $\mathcal{M}_u$ is to behave indistinguishably from the retrained model $\mathcal{M}_r$.  
Consequently, many evaluation approaches rely on comparisons with the retrained model. 
However, for LLMs, complete retraining is generally infeasible due to the scale of the training data and the inaccessibility of proprietary datasets to external auditors.

Thus, the retrained model cannot serve as a direct reference in the LLM setting.  
However, by extrapolating from the principles underlying $\mathcal{M}_r$, we can identify the core objectives for unlearning:
\begin{quote}
%\label{goal}
    \em \textbf{Goal of LLM unlearning}: the unlearned model should no longer memorize content from the unlearn set while preserving all other content.
\end{quote} 
Meanwhile, we expect the unlearning algorithm to achieve the above objectives with minimal computational and time overhead.
Guided by these goals, numerous studies have proposed corresponding evaluation metrics, which we examine in detail in Section~\ref{sec:eval}.

\subsection{Related Topics}
Several related research areas exhibit conceptual or methodological overlaps with LLM unlearning, offering valuable insights and transferable techniques. 
However, their core objectives and problem formulations differ from the LLM unlearning paradigm. 
Hence, we briefly introduce these adjacent fields to clarify correlations and distinctions in this section, while detailed discussions of these topics fall beyond this survey's scope.

\subsubsection{Memorization and Data Extraction}
As formalized by the goal of unlearning in Section~\ref{sec:def_goal}, the conceptualization of memorization directly shapes the objectives of unlearning, while the methodology for memorization detection provides an essential diagnostic tool for evaluating unlearning efficacy. 
There exist multiple definitions of LLM memorization, such as formulations based on counterfactual memorization~\citep{zhang2023counterfactual} and tuple completion~\citep{meng2022locating}. 
Among these, extractable memorization~\citep{carlini2021extracting} is the most prevalent, conceptualizing memorization as content that the model can reproduce under specific prompting conditions. 
This definition originally involved identifying a precise input prefix to induce the model to output the memorized content, and has evolved into a diverse class of data extraction attacks, employing various input strategies and detection mechanisms~\citep{shin2020autoprompt,zou2023universal}. 
Consequently, data extraction attacks serve a dual role: they constitute a critical tool for evaluating unlearning, particularly for assessing the knowledge memorization, while unlearning itself functions as a defensive measure to purge hazardous knowledge and thereby mitigate the risks posed by malicious data extraction attempts.

% As described in the ``knowledge-level goal'' in Section~\ref{sec:def_goal}, the definition of memorization directly affects the goal of unlearning, and the discrimination method of memorization provides a necessary diagnostic tool for the evaluation of unlearning.
% There are many ways to define the LLM memorization, such as counterfactual memorization~\citep{zhang2023counterfactual}, tuple completion~\citep{meng2022locating}.
% Among them, extractable memorization~\citep{carlini2021extracting} is the most commonly used, whose core idea regards memorization as the content that the model can output under specific conditions.
% This definition initially requires finding a specific prefix input to induce model output, and later evolved into a rich set of data extraction attacks with various forms of inputs and detection methods~\citep{shin2020autoprompt,zou2023universal}.
% On the one hand, data extraction attacks play an important role in unlearning evaluation, especially in the evaluation of knowledge memorization. 
% On the other hand, unlearning helps to eliminate unsafe knowledge from the model to defend against malicious data extraction attacks.

\subsubsection{Knowledge Updating}
Knowledge editing and updating are essential for maintaining the long-term efficacy of large language models (LLMs), as they enable the correction of inaccuracies and the integration of new knowledge without requiring full model retraining. 
LLM unlearning can be viewed as a promising strategy within this domain, with research advancing in two main directions: some studies develop robust, conflict-free parameter update algorithms to facilitate reliable knowledge modification~\citep{jung2025come,ni2024forgetting,sun2024learning}, while others apply unlearning techniques to domain-specific contexts~\citep{yang2024hotfixing,fore2024unlearning}. 
Another widely adopted paradigm is model editing, which focuses on local, targeted modifications to specific factual knowledge while preserving the model's general capabilities and avoiding catastrophic forgetting. 
A key distinction between model editing and unlearning lies in their objectives: model editing operates with a predefined target knowledge state, whereas unlearning aims to remove or suppress information without necessarily replacing it. 
Nevertheless, mechanistic insights from model editing techniques, such as knowledge neurons and locate-then-edit approaches~\citep{meng2022locating}, can inform the design of more precise and interpretable unlearning methods.

% Knowledge editing and updating, which allows LLM to correct errors or integrate new information without the need for complete retraining, is crucial to continuous effectiveness of LLMs over time.
% LLM unlearning has also shown great promise in knowledge editing and updating, with some studies proposing robust and conflict-free parameter updating algorithms~\citep{jung2025come,ni2024forgetting,sun2024learning} while others focusing on specific fields~\citep{yang2024hotfixing,fore2024unlearning}.
% Another commonly used approach is model editing, which locally modify specific factual knowledge of LLM without catastrophic forgetting of general capabilities.
% Compared to LLM unlearning, model editing updates model with a clear target knowledge, which unlearning does not have. 
% Nevertheless, some techniques in model editing like knowledge neuron and locate-then-edit~\citep{meng2022locating} can inform the mechanism design for unlearning methods.

\subsubsection{Alignment}
Alignment seeks to ensure that LLMs behave in accordance with human values and intentions.
This objective is dual in nature, involving the guide of models toward generating helpful responses (positive) and preventing them from producing undesirable outputs (negative). 
A widely adopted approach for positive guidance is reinforcement learning from human feedback (RLHF), which steers models toward desirable behaviors through iterative reward-based optimization~\citep{bai2022training,ouyang2022training}. 
Complementarily, LLM unlearning has emerged as a critical technique for negative alignment, systematically removing undesirable knowledge or capabilities from models. 
For example, it has been applied to mitigate social biases~\citep{Yu2023Unlearning,Dige2024Mitigating}, eliminate unauthorized content to protect copyright~\citep{Wei2024Evaluating}, and reduce the risk of leaking sensitive information~\citep{Rashid2024Forget,geng2025mitigating}. 
Together, these methods form a cohesive alignment framework that addresses both the promotion of beneficial behaviors and the suppression of harmful ones.

% Alignment aims to guide LLMs in their behavior in accordance with human values and intentions.
% The goal of alignment is twofold. 
% Reinforcement learning from human feedback (RLHF) is a commonly used method to guide the model to do the right thing (e.g., providing useful information).
% Meanwhile, LLM unlearning is an important alignment method to prevent the model from doing the wrong or harmful thing, which has been employed to reduce social biases~\citep{Yu2023Unlearning,Dige2024Mitigating}, remove unauthorized content to protect copyright~\citep{Wei2024Evaluating}, and mitigate the risk of sensitive information leakage~\citep{Rashid2024Forget,geng2025mitigating}.
% !TeX root = ../main_csur.tex
\section{Existing unlearning methods}
\label{sec:methods}
% !TeX root = ../main_csur.tex
\begin{figure}[t]
    \centering
    \scriptsize
    \begin{tikzpicture}
        \node (origin) at (0,2.6) [rectangle,draw=black,line width=1pt,rounded corners=5pt, minimum width=60pt] {\footnotesize \makecell{Original model,\\data, etc.}};
        \node (output) at (0,-2.4) [rectangle,draw=black,line width=1pt,rounded corners=5pt, minimum width=60pt] {\footnotesize Output};
        \node (base) at (0,0) [rectangle,draw=black,line width=1pt,rounded corners=5pt, text width=50pt, inner sep=5pt, align=center] {\includegraphics[width=1cm]{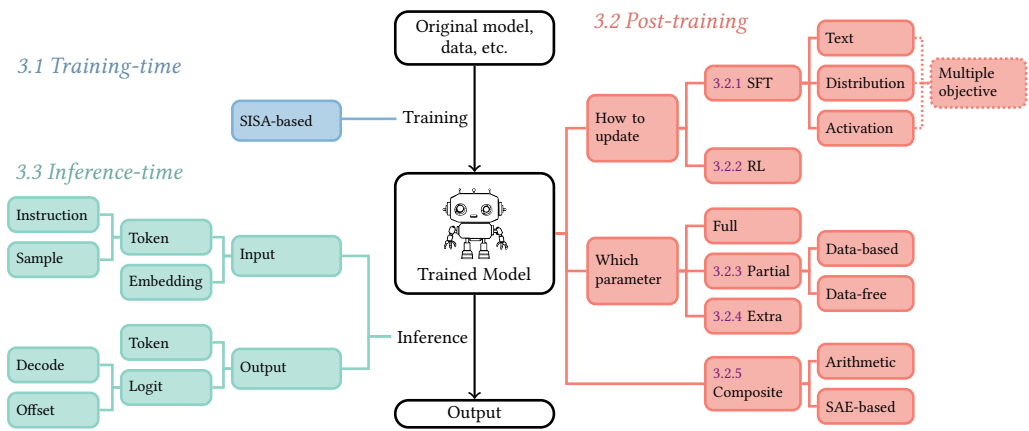}\\\footnotesize Trained Model};
        \draw [->, thick] (origin.south) -- node(train)[left,draw=none,fill=none] {\footnotesize Training} (base.north);
        \draw [->, thick] (base.south) -- node(infer)[yshift=4pt,left,draw=none,fill=none] {\footnotesize Inference} (output.north);

        \node (sisa) at (-2.5,1.5) [rectangle, draw=blue, fill=lightblue, text width=35pt, inner sep=3pt, minimum height=15pt, align=left, rounded corners=3pt, line width=1pt] {SISA-based};
        %\node (incontext) at (-4,1.6) [rectangle, draw=blue, fill=lightblue, text width=30pt, inner sep=3pt, minimum height=13pt, align=left, rounded corners=3pt, line width=1pt] {In-context};
        %\draw [-,thick,draw=blue,line width=1pt] (train.west) --++ (-1.8,0) |- (sisa.east);
        \draw [-,thick,draw=blue,line width=1pt] (train.west) --++ (-0.75,0);
        %\draw [-,thick,draw=blue,line width=1pt] (train.west) -- node []{\tiny\textcolor{blue!80!black}{\makecell{Learning\\architecture}}} ++ (-1.8,0) |- (incontext.east);
        \node at (-5,2.2) [rectangle,fill=white,inner sep=2pt] {\textcolor{blue!80!black}{\normalsize \em 3.1 Training-time}};

        %%%%%%%%%%%%%%%%%%%%%%
        
        \node (input) at (-2.5,-0.3) [rectangle, draw=green, fill=lightgreen, text width=35pt, inner sep=3pt, minimum height=15pt, align=left, rounded corners=3pt, line width=1pt] {Input};
        \node (output) at (-2.5,-1.8) [rectangle, draw=green, fill=lightgreen, text width=35pt, inner sep=3pt, minimum height=15pt, align=left, rounded corners=3pt, line width=1pt] {Output};
        \draw [-,thick,draw=green,line width=1pt] (infer.west) --++ (-0.3,0) |- (input.east);
        \draw [-,thick,draw=green,line width=1pt] (infer.west) --++ (-0.3,0) |- (output.east);
        %\draw [-,thick,draw=green,line width=1pt] (infer.west) -- node []{\tiny\textcolor{green!80!black}{\makecell{Modify\\what}}} ++ (-0.8,0) |- (output.east);
        
        \node (in_token) at (-4.1,-0.05) [rectangle, draw=green, fill=lightgreen, text width=28pt, inner sep=3pt, minimum height=13pt, align=left, rounded corners=3pt, line width=1pt] {Token};
        \node (in_embed) at (-4.1,-0.65) [rectangle, draw=green, fill=lightgreen, text width=28pt, inner sep=3pt, minimum height=13pt, align=left, rounded corners=3pt, line width=1pt] {Embedding};
        \draw [-,thick,draw=green,line width=1pt] (input.west) --++ (-0.1,0) |- (in_token.east);
        \draw [-,thick,draw=green,line width=1pt] (input.west) --++ (-0.1,0) |- (in_embed.east);
        
        \node (out_token) at (-4.1,-1.45) [rectangle, draw=green, fill=lightgreen, text width=28pt, inner sep=3pt, minimum height=13pt, align=left, rounded corners=3pt, line width=1pt] {Token};
        \node (out_logit) at (-4.1,-2.05) [rectangle, draw=green, fill=lightgreen, text width=28pt, inner sep=3pt, minimum height=13pt, align=left, rounded corners=3pt, line width=1pt] {Logit};
        \draw [-,thick,draw=green,line width=1pt] (output.west) --++ (-0.1,0) |- (out_token.east);
        \draw [-,thick,draw=green,line width=1pt] (output.west) --++ (-0.1,0) |- (out_logit.east);

        \node (instruct) at (-5.6,0.25) [rectangle, draw=green, fill=lightgreen, text width=28pt, inner sep=3pt, minimum height=13pt, align=left, rounded corners=3pt, line width=1pt] {Instruction};
        \node (sample) at (-5.6,-0.35) [rectangle, draw=green, fill=lightgreen, text width=28pt, inner sep=3pt, minimum height=13pt, align=left, rounded corners=3pt, line width=1pt] {Sample};
        \draw [-,thick,draw=green,line width=1pt] (in_token.west) --++ (-0.1,0) |- (instruct.east);
        \draw [-,thick,draw=green,line width=1pt] (in_token.west) --++ (-0.1,0) |- (sample.east);
        %\node (sam_sel) at (-5.5,-0.7) [rectangle, draw=green, fill=lightgreen, text width=30pt, inner sep=3pt, minimum height=13pt, align=left, rounded corners=3pt, line width=1pt] {Sample select};

        \node (decode) at (-5.6,-1.75) [rectangle, draw=green, fill=lightgreen, text width=28pt, inner sep=3pt, minimum height=13pt, align=left, rounded corners=3pt, line width=1pt] {Decode};
        \node (offset) at (-5.6,-2.35) [rectangle, draw=green, fill=lightgreen, text width=28pt, inner sep=3pt, minimum height=13pt, align=left, rounded corners=3pt, line width=1pt] {Offset};
        \draw [-,thick,draw=green,line width=1pt] (out_logit.west) --++ (-0.1,0) |- (decode.east);
        \draw [-,thick,draw=green,line width=1pt] (out_logit.west) --++ (-0.1,0) |- (offset.east);
        
        \node at (-5,0.8) [rectangle,fill=white,inner sep=2pt] {\textcolor{green!80!black}{\normalsize \em 3.3 Inference-time}};

        %%%%%%%%%%%%%%%%%%%%%%

        \node (ft) at (2.1,1.4) [rectangle, draw=red, fill=lightred, text width=28pt, inner sep=3pt, minimum height=25pt, align=left, rounded corners=3pt, line width=1pt] {\makecell[l]{How to\\update}};
        \node (params) at (2.1,-0.5) [rectangle, draw=red, fill=lightred, text width=28pt, inner sep=3pt, minimum height=25pt, align=left, rounded corners=3pt, line width=1pt] {\makecell[l]{Which\\parameter}};
        \draw [-,thick,draw=red,line width=1pt] (base.east) --++ (0.15,0) |- (ft.west);
        \draw [-,thick,draw=red,line width=1pt] (base.east) --++ (0.15,0) |- (params.west);
        
        \node (sft) at (3.7,2) [rectangle, draw=red, fill=lightred, text width=30pt, inner sep=3pt, minimum height=13pt, align=left, rounded corners=3pt, line width=1pt] {\ref{sec:ft_sft} SFT};
        \node (rl) at (3.7,0.9) [rectangle, draw=red, fill=lightred, text width=30pt, inner sep=3pt, minimum height=13pt, align=left, rounded corners=3pt, line width=1pt] {\ref{sec:ft_rl} RL};
        \draw [-,thick,draw=red,line width=1pt] (ft.east) --++ (0.1,0) |- (sft.west);
        \draw [-,thick,draw=red,line width=1pt] (ft.east) --++ (0.1,0) |- (rl.west);

        \node (text) at (5.2,2.6) [rectangle, draw=red, fill=lightred, text width=30pt, inner sep=3pt, minimum height=13pt, align=left, rounded corners=3pt, line width=1pt] {Text};
        \node (distr) at (5.2,2) [rectangle, draw=red, fill=lightred, text width=30pt, inner sep=3pt, minimum height=13pt, align=left, rounded corners=3pt, line width=1pt] {Distribution};
        \node (activ) at (5.2,1.4) [rectangle, draw=red, fill=lightred, text width=30pt, inner sep=3pt, minimum height=13pt, align=left, rounded corners=3pt, line width=1pt] {Activation};
        \draw [-,thick,draw=red,line width=1pt] (sft.east) --++ (0.1,0) |- (text.west);
        \draw [-,thick,draw=red,line width=1pt] (sft.east) --++ (0.1,0) |- (distr.west);
        \draw [-,thick,draw=red,line width=1pt] (sft.east) --++ (0.1,0) |- (activ.west);

        \node (multi) at (6.7,2) [rectangle, draw=red, densely dotted, fill=lightred, text width=30pt, inner sep=3pt, minimum height=13pt, align=left, rounded corners=3pt, line width=1pt] {Multiple objective};
        \draw [-,thick,densely dotted,draw=red,line width=1pt] (text.east) --++ (0.1,0) |- (multi.west);
        \draw [-,thick,densely dotted,draw=red,line width=1pt] (distr.east) --++ (0.1,0) |- (multi.west);
        \draw [-,thick,densely dotted,draw=red,line width=1pt] (activ.east) --++ (0.1,0) |- (multi.west);
        %\draw [-,thick,draw=red,line width=1pt] (method.east) -- node []{\tiny\textcolor{red!80!black}{\makecell{Based\\on}}} ++ (0.1,0) |- (activ.west);

        \node (full) at (3.7,0.1) [rectangle, draw=red, fill=lightred, text width=30pt, inner sep=3pt, minimum height=13pt, align=left, rounded corners=3pt, line width=1pt] {Full};
        \node (partial) at (3.7,-0.5) [rectangle, draw=red, fill=lightred, text width=30pt, inner sep=3pt, minimum height=13pt, align=left, rounded corners=3pt, line width=1pt] {\ref{sec:ft_local} Partial};
        \node (extra) at (3.7,-1.1) [rectangle, draw=red, fill=lightred, text width=30pt, inner sep=3pt, minimum height=13pt, align=left, rounded corners=3pt, line width=1pt] {\ref{sec:ft_new} Extra};
        \draw [-,thick,draw=red,line width=1pt] (params.east) --++ (0.1,0) |- (full.west);
        \draw [-,thick,draw=red,line width=1pt] (params.east) --++ (0.1,0) |- (partial.west);
        \draw [-,thick,draw=red,line width=1pt] (params.east) --++ (0.1,0) |- (extra.west);

        \node (par_data) at (5.2,-0.2) [rectangle, draw=red, fill=lightred, text width=30pt, inner sep=3pt, minimum height=13pt, align=left, rounded corners=3pt, line width=1pt] {Data-based};
        \node (par_no_data) at (5.2,-0.8) [rectangle, draw=red, fill=lightred, text width=30pt, inner sep=3pt, minimum height=13pt, align=left, rounded corners=3pt, line width=1pt] {Data-free};
        \draw [-,thick,draw=red,line width=1pt] (partial.east) --++ (0.1,0) |- (par_data.west);
        \draw [-,thick,draw=red,line width=1pt] (partial.east) --++ (0.1,0) |- (par_no_data.west);

        \node (comp) at (3.7,-2) [rectangle, draw=red, fill=lightred, text width=30pt, inner sep=3pt, minimum height=13pt, align=left, rounded corners=3pt, line width=1pt] {\makecell[l]{\ref{sec:ft_others}\\ Composite}};
        \draw [-,thick,draw=red,line width=1pt] (base.east) --++ (0.15,0) |- (comp.west);
        %\node at (3.2,-1.6) {\ref{sec:ft_others} Composite Approaches};
        \node (task_vec) at (5.2,-1.7) [rectangle, draw=red, fill=lightred, text width=30pt, inner sep=3pt, minimum height=13pt, align=left, rounded corners=3pt, line width=1pt] {Arithmetic};
        \node (subspace) at (5.2,-2.3) [rectangle, draw=red, fill=lightred, text width=30pt, inner sep=3pt, minimum height=13pt, align=left, rounded corners=3pt, line width=1pt] {SAE-based};
        \draw [-,thick,draw=red,line width=1pt] (comp.east) --++ (0.1,0) |- (task_vec.west);
        \draw [-,thick,draw=red,line width=1pt] (comp.east) --++ (0.1,0) |- (subspace.west);

        % \node (task_vec) at (3.7,-1.15) [rectangle, draw=red, fill=lightred, text width=30pt, inner sep=3pt, minimum height=13pt, align=left, rounded corners=3pt, line width=1pt] {Arithmetic operations};
        % \node (subspace) at (3.7,-1.9) [rectangle, draw=red, fill=lightred, text width=30pt, inner sep=3pt, minimum height=13pt, align=left, rounded corners=3pt, line width=1pt] {SAE-based};
        % \node (etc) at (3.7,-2.5) [rectangle, draw=red, fill=lightred, text width=30pt, inner sep=3pt, minimum height=13pt, align=left, rounded corners=3pt, line width=1pt] {...};
        % \draw [-,thick,draw=red,line width=1pt] (other.east) --++ (0.1,0) |- (task_vec.west);
        % \draw [-,thick,draw=red,line width=1pt] (other.east) --++ (0.1,0) |- (subspace.west);
        % \draw [-,thick,draw=red,line width=1pt] (other.east) --++ (0.1,0) |- (etc.west);

        % \node (peft) at (5.2,-0.9) [rectangle, draw=red, fill=lightred, text width=30pt, inner sep=3pt, minimum height=13pt, align=left, rounded corners=3pt, line width=1pt] {PEFT};
        % \node (subspace) at (5.2,-1.5) [rectangle, draw=red, fill=lightred, text width=30pt, inner sep=3pt, minimum height=13pt, align=left, rounded corners=3pt, line width=1pt] {Subspace};
        % \draw [-,thick,draw=red,line width=1pt] (task_vec.east) --++ (0.1,0) |- (peft.west);
        % \draw [-,thick,draw=red,line width=1pt] (task_vec.east) --++ (0.1,0) |- (subspace.west);
        
        \node at (2.6,2.8) [rectangle,fill=none,inner sep=2pt] {\textcolor{red!90!black}{\normalsize \em 3.2 Post-training}};

    \end{tikzpicture}
    \caption{Framework of unlearning methods. In typical LLM usage scenarios, a model is first trained on specific datasets, and then is used for inference to generate outputs. The unlearning method can be applied to the training process, the trained model, or the inference stage, corresponding to training-time unlearning (Section \ref{sec:training-time}, post-training unlearning (Section \ref{sec:post-training}) and inference-time unlearning (Section \ref{sec:infer-time}).}
    \label{fig:method_frame}
\end{figure}

In typical LLM usage scenarios, a model is first trained on specific datasets from draft or a pretrained base model, and then is used for inference to generate output in some tasks. 
As illustrated in Figure~\ref{fig:method_frame}, the unlearning method can be applied to the training process, the trained model, or the inference stage, corresponding to training-time unlearning (Section \ref{sec:training-time}), post-training unlearning (Section \ref{sec:post-training}) and inference-time unlearning (Section \ref{sec:infer-time}).
%We categorize most existing LLM Unlearning methods into three main types according to the different stages of training, fine-tuning, and inference of LLM in which the unlearning algorithm participates: Training-Time Unlearning (\S~\ref{sec:training-time}), Post-Training Unlearning (\S~\ref{sec:post-training}), and Inference-Time Unlearning (\S~\ref{sec:infer-time}). 

In short, \textbf{Training-Time Unlearning} requires adjusting the training process to facilitate unlearning, which is mainly based on SISA training paradigms. 
\textbf{Post-Training Unlearning} involves altering the trained model, mainly through supervised fine-tuning (SFT) or reinforcement training (RL) on selected parameters. 
\textbf{Inference-time Unlearning} aims to achieve unlearning via input or output adjustments, rather than modifying the model parameters.

\subsection{Training-Time Unlearning}
\label{sec:training-time}
As the pretraining of LLMs typically involves complex procedures and massive datasets, existing training-time unlearning methods primarily focus on the further training phase of a pretrained base model. 
These approaches address a setting in which a general base model $\mathcal{M}_b$ is adapted for specific downstream tasks, during which it may have memorized sensitive information and thus requires unlearning. 
%Based on the adaptation strategy employed, these methods can be classified into two categories: (1) parameter fine-tuning and (2) in-context learning.

%In the case of parameter fine-tuning, a
As noted in Section~\ref{sec:def}, the ideal outcome of unlearning is to obtain a retrained model $\mathcal{M}_r$. 
However, full retraining starting from $\mathcal{M}_b$ is computationally prohibitive and impractical in real-world scenarios. 
To alleviate this burden, training-time unlearning techniques, exemplified by SISA~\citep{Bourtoule2021Machine}, introduce novel training frameworks that initiate retraining from intermediate states. 
These methods partition the dataset into multiple subsets and store corresponding checkpoints trained on different subsets. 
By constraining the influence of data points, retraining can start from specific checkpoints unaffected by the unlearn set, thereby accelerating the unlearning process. 

Given the considerable storage and computational overhead associated with maintaining numerous model copies, \citet{BannihattiKumar2023Privacy} and \citet{Chowdhury2024Scalable} integrate supplementary trainable components, applying the SISA principle to fine-tune and preserve only the newly introduced parameters. 
This strategy substantially reduces the number of parameters requiring updates. 
Beyond efficiency and performance considerations, \citet{kadhe2023fairsisa} examine fairness concerns in SISA-based frameworks and propose FairSISA, which integrates three post-processing bias mitigation techniques.

In summary, training-time unlearning ensures, from a mechanistic standpoint, that the model does not encounter data from the unlearn, thereby providing verifiable guarantees. 
However, this approach is inapplicable to models that have already been fully trained, which significantly constrains its practical applicability.

\subsection{Post-Training Unlearning}
\label{sec:post-training}
The main methodology for unlearning in LLMs involves adjusting the parameters of a pre-trained model, a phase typically termed ``post-training''. 
This approach raises two fundamental questions:
\begin{quote}
    \em
    (Q1) How should parameters be modified?\\
    (Q2) Which parameters should be targeted for modification?
\end{quote}
To address (Q1), following the conventional taxonomy of LLM training, we categorize the unlearning methods into supervised fine-tuning (SFT) and reinforcement learning (RL).
Regarding (Q2), the scope of parameter modification may encompass the entire model, a selected subset of parameters, or newly introduced parameters.
Strategies for selecting a subset of parameters will be discussed in Section \ref{sec:ft_local}, while methods for integrating new parameters are examined in Section \ref{sec:ft_new}.
It is important to note that parameter selection strategies and training algorithms are orthogonal and can be combined flexibly.
As a complementary perspective, Section~\ref{sec:ft_others} reviews notable hybrid approaches that integrate multiple techniques discussed in the preceding subsections.

% The main approach to unlearning is to modify the parameters of trained LLMs, which is commonly referred to as the ``post-training'' phase. 
% %Based on the scope of parameter modification, unlearning methods in this category can be further divided into three types: \textbf{full parameter updating}, \textbf{partial parameter updating}, and \textbf{importing extra parameters}. We will introduce these three types in the following subsections.
% This leads to two crucial questions:
% (1) How to modify the parameters? 
% (2) Which parameters should be selected for modification?
% For question (1), following the usual classification of LLM training, we divide methods into supervised fine-tuning (SFT) and reinforcement learning (RL).
% %the modification of parameters in most methods is an optimization problem, so we will introduce their objective design in Section \ref{sec:ft_sft}.
% %While some methods adopt alternative strategies, we include important ones in Section \ref{sec:ft_others}.
% For question (2), the modified parameters can be all parameters of the model, part of the parameters, or newly introduced parameters.
% We will discuss how to select part of the parameters in Section \ref{sec:ft_local} and how to incorporate new parameters in Section \ref{sec:ft_new}.
% Note that parameter strategies can be freely combined with the training algorithms.
% As a complementary perspective, we review some notable approaches that combine several techniques in the first four sections in Section~\ref{sec:ft_others}.

\subsubsection{Supervised Fine-tuning (SFT)}
\label{sec:ft_sft}
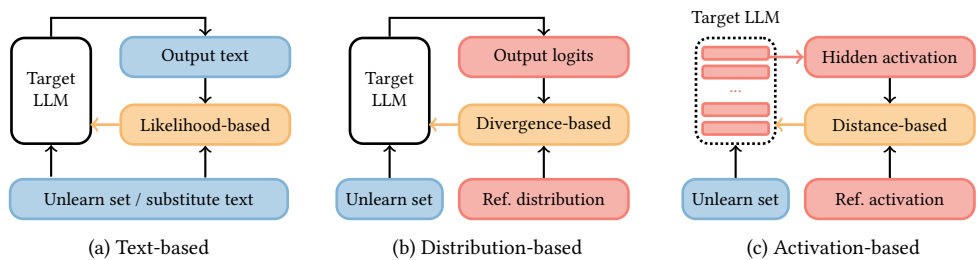
\begin{figure}[t]
    \footnotesize
    \centering
    % \begin{subfigure}{0.32\textwidth}
    %     \centering
    %     \includegraphics[width=\textwidth]{figs/sec3/text-based.pdf}
    %     \caption{Text-based}
    %     \label{fig:sub1}
    % \end{subfigure}
    % % \hfill
    % \begin{subfigure}{0.32\textwidth}
    %     \centering
    %     \includegraphics[width=\textwidth]{figs/sec3/dist-based.pdf}
    %     \caption{Distribution-based}
    %     \label{fig:sub2}
    % \end{subfigure}
    % % \hfill
    % \begin{subfigure}{0.32\textwidth}
    %     \centering
    %     \includegraphics[width=\textwidth]{figs/sec3/act-based.pdf}
    %     \caption{Activation-based}
    %     \label{fig:sub3}
    % \end{subfigure}
    \begin{tikzpicture}
        \node (target1) at (-4.5,0) [rectangle, draw=black, fill=none, minimum width=30pt, minimum height=40pt, inner sep=3pt, rounded corners=5pt, line width=1pt] {\makecell{Target\\LLM}};
        \node (pre_log1) at (-2.45,0.45) [rectangle, draw=blue, fill=lightblue, minimum width=64pt, minimum height=15pt, inner sep=3pt, rounded corners=5pt, line width=1pt] {Output text};
        %\node (tok_label) at (-2.7,-1) [rectangle, draw=red, fill=lightred, minimum width=70pt, minimum height=15pt, inner sep=3pt, rounded corners=5pt, line width=1pt] {Token-wise label};
        \node (loss1) at (-2.45,-0.45) [rectangle, draw=orange, fill=lightorange, minimum width=64pt, minimum height=15pt, inner sep=3pt, rounded corners=5pt, line width=1pt] {{Likelihood-based}};
        \node (input1) at (-3.2,-1.44) [rectangle, draw=blue, fill=lightblue, minimum width=105pt, minimum height=15pt, inner sep=3pt, rounded corners=5pt, line width=1pt] {Unlearn set / substitute text};
        \draw [->, thick, draw=black] ([yshift=-12pt]target1.south) -- (target1.south);
        \draw [->, thick, draw=black] (target1.north) --++(0,0.25) -| (pre_log1.north);
        \draw [->, thick, draw=black] (pre_log1.south) -- (loss1.north);
        \draw [->, thick, draw=black] ([yshift=-12pt]loss1.south) -- (loss1.south);
        %\draw [->, thick, draw=black] (tok_label.north) -- (loss1.south);
        \draw [->, thick, draw=orange, line width=1pt] (loss1.west) -- ([xshift=-11pt]loss1.west);
        \node at (-3.2,-2.1) {\small (a) Text-based};

        \node (target2) at (0,0) [rectangle, draw=black, fill=none, minimum width=30pt, minimum height=40pt, inner sep=3pt, rounded corners=5pt, line width=1pt] {\makecell{Target\\LLM}};
        \node (pre_log2) at (2.05,0.45) [rectangle, draw=red, fill=lightred, minimum width=64pt, minimum height=15pt, inner sep=3pt, rounded corners=5pt, line width=1pt] {Output logits};
        \node (ref_log) at (2.05,-1.44) [rectangle, draw=red, fill=lightred, minimum width=64pt, minimum height=15pt, inner sep=3pt, rounded corners=5pt, line width=1pt] {Ref. distribution};
        \node (loss2) at (2.05,-0.45) [rectangle, draw=orange, fill=lightorange, minimum width=64pt, minimum height=15pt, inner sep=3pt, rounded corners=5pt, line width=1pt] {{Divergence-based}};
        \node (input2) at (0,-1.44) [rectangle, draw=blue, fill=lightblue, minimum width=40pt, minimum height=15pt, inner sep=3pt, rounded corners=5pt, line width=1pt] {Unlearn set};
        %\node (input2-2) at (2.3,-2) [rectangle, draw=blue, fill=lightblue, minimum width=70pt, minimum height=15pt, inner sep=3pt, rounded corners=5pt, line width=1pt] {Reference distribution};
        \draw [->, thick, draw=black] (input2) -- (target2.south);
        \draw [->, thick, draw=black] (target2.north) --++(0,0.25) -| (pre_log2.north);
        \draw [->, thick, draw=black] (pre_log2.south) -- (loss2.north);
        %\draw [->, thick, draw=black] (input2-2) -- (ref_log.south);
        \draw [->, thick, draw=black] (ref_log.north) -- (loss2.south);
        \draw [->, thick, draw=orange, line width=1pt] (loss2.west) -- ([xshift=-11pt]loss2.west);
        \node at (1.3,-2.1) {\small (b) Distribution-based};

        \node at (4.6,0.5) [rectangle, draw=red, fill=lightred, minimum width=25pt, minimum height=1pt, rounded corners=1pt, line width=1pt] {};
        \node at (4.6,0.25) [rectangle, draw=red, fill=lightred, minimum width=25pt, minimum height=1pt, rounded corners=1pt, line width=1pt] {};
        \node at (4.6,-0.25) [rectangle, draw=red, fill=lightred, minimum width=25pt, minimum height=1pt, rounded corners=1pt, line width=1pt] {};
        \node at (4.6,-0.5) [rectangle, draw=red, fill=lightred, minimum width=25pt, minimum height=1pt, rounded corners=1pt, line width=1pt] {};
        \node (target3) at (4.6,0) [rectangle, densely dotted, draw=black, fill=none, minimum width=30pt, minimum height=40pt, inner sep=3pt, rounded corners=5pt, line width=1pt] {\textcolor{red!90!black}{...}};
        \node at (4.6,0.95) {Target LLM};

        \node (act) at (6.65,0.45) [rectangle, draw=red, fill=lightred, minimum width=64pt, minimum height=15pt, inner sep=3pt, rounded corners=5pt, line width=1pt] {Hidden activation};
        \node (ref_act) at (6.65,-1.44) [rectangle, draw=red, fill=lightred, minimum width=64pt, minimum height=15pt, inner sep=3pt, rounded corners=5pt, line width=1pt] {Ref. activation};
        \node (loss3) at (6.65,-0.45) [rectangle, draw=orange, fill=lightorange, minimum width=64pt, minimum height=15pt, inner sep=3pt, rounded corners=5pt, line width=1pt] {{Distance-based}};
        \node (input3) at (4.6,-1.44) [rectangle, draw=blue, fill=lightblue, minimum width=40pt, minimum height=15pt, inner sep=3pt, rounded corners=5pt, line width=1pt] {Unlearn set};
        \draw [->, thick, draw=black] (input3) -- (target3.south);
        \draw [->, thick, draw=red,line width=1pt] ([xshift=-14pt]act.west) -- (act.west);
        \draw [->, thick, draw=black] (act.south) -- (loss3.north);
        %\draw [->, thick, draw=black] (input2-2) -- (ref_log.south);
        \draw [->, thick, draw=black] (ref_act.north) -- (loss3.south);
        \draw [->, thick, draw=orange,line width=1pt] (loss3.west) -- ([xshift=-11pt]loss3.west);
        \node at (5.9,-2.1) {\small (c) Activation-based};

        % \node (pre_act) at (2.3,1) [rectangle, draw=red, fill=lightred, minimum width=70pt, minimum height=15pt, inner sep=3pt, rounded corners=5pt, line width=1pt] {Prediction logits};
        % \node (ref_log) at (2.3,-1) [rectangle, draw=red, fill=lightred, minimum width=70pt, minimum height=15pt, inner sep=3pt, rounded corners=5pt, line width=1pt] {Reference logits};
        % \node (loss2) at (2.3,0) [rectangle, draw=orange, fill=lightorange, minimum width=70pt, minimum height=15pt, inner sep=3pt, rounded corners=5pt, line width=1pt] {\textbf{Divergence-based}};
        % \node (input2) at (0,-2) [rectangle, draw=blue, fill=lightblue, minimum width=40pt, minimum height=15pt, inner sep=3pt, rounded corners=5pt, line width=1pt] {Unlearn set text};
        % \node (input2-2) at (2.3,-2) [rectangle, draw=blue, fill=lightblue, minimum width=70pt, minimum height=15pt, inner sep=3pt, rounded corners=5pt, line width=1pt] {Reference distribution};
        % \draw [->, thick, draw=black] (input2) -- (target2.south);
        % \draw [->, thick, draw=black] (target2.north) --++(0,0.2) -| (pre_log2.north);
        % \draw [->, thick, draw=black] (pre_log2.south) -- (loss2.north);
        % \draw [->, thick, draw=black] (input2-2) -- (ref_log.south);
        % \draw [->, thick, draw=black] (ref_log.north) -- (loss2.south);
        % \draw [->, thick, draw=black] (loss2.west) -- ([xshift=-14pt]loss2.west);
    \end{tikzpicture}
    \caption{
    Objective designs of unlearning methods. 
    The color coding is as follows: \textcolor{blue!90!black}{blue for text,} \textcolor{red!90!black}{red for tensors/vectors,} \textcolor{orange!90!black}{orange for loss functions}. 
    Text-based and distribution-based methods compute a loss function at the output layer by comparing it to a reference (ref.), in textual and distributional level, respectively.
    Activation-based methods compute the loss using activations from the hidden layers against a reference.
    % Illustration of different categories on objective design. \textcolor{blue!90!black}{Blue block} denotes the text. \textcolor{red!90!black}{Red block} denotes the tensor or vector. \textcolor{orange!90!black}{Orange block} denotes the loss function. Text-based and distribution-based methods calculate a loss on the output of the model with a reference (ref.), where the former is based on text and the latter is based on distribution. Activation-based methods calculate a loss on the activation of hidden layers in the model with a reference (ref.).
    }
    \label{fig:loss-design}
\end{figure}

Supervised Fine-Tuning (SFT) adapts an LLM to downstream tasks by training it on labeled, task-specific data.
The core of an SFT-based unlearning mechanism is the design of its objective for optimization, which dictates how the model's parameters are adjusted to forget specific knowledge while minimizing the impact on general utility.
Different from the unified next-token prediction loss during learning, the design of unlearning objective is quite more diverse.
Based on the primary target of the designated objective, we can classify existing methods into three major categories: \textbf{Text-based, Distribution-based, and Activation-based}.
Figure~\ref{fig:loss-design} compares the general pipeline of different objective categories.

\textbf{Text-based}. 
These kind of objectives are most intuitive, which aim at minimize or maximize the predicted likelihood of certain text. 
A representative baseline is Gradient Ascent~\cite{jang2022knowledge, Yao2024Large}. 
It reduces the prediction probabilities of forget set samples by directly negating their cross-entropy next-token prediction loss.
% \begin{equation*}
%     \mathcal{L}_{\mathrm{GA}} = -\mathbb{E}_{(x,y)\thicksim\mathcal{D}_\mathrm{F}}\left[-\log p(y|x;\theta)\right]
% \end{equation*}
Some objectives also draw inspiration from preference optimization: 
NPO~\cite{Zhang2024Negative} introduces a reference model to constrain parameter changes, and SimNPO~\cite{fan2025simplicity} further improves NPO with length-normalized reward.
Another approach focuses on improving likelihood of ``substitute responses'', which gives appropriate reply to queries from forget set, without disclosing targeted knowledge~\cite{Maini2024TOFU, Mekala2024Alternate}. 
However, these simple baselines often impair general ability of unlearned model, as unlearning corpus typically includes a substantial proportion of tokens that are irrelevant to target knowledge.
To alleviate this, some propose to choose only key part in forget set sequences for optimization~\cite{Wang2024Selective} or apply different weights to various token positions within the sequences of forget set~\cite{Wang2024Rethinking, Feng2024Finegrained}.
More methods put attention on generating or selecting data. Some utilize an external LLM to generate substitute responses relevant to forget set queries~\cite{Mekala2024Alternate, Sinha2024UnStar, xu2025relearn, Liu2024Learning, Gu2024MEOW}.
\citet{Patil2025UPCORE} and \citet{Chang2025Whicha} proposed methods for selecting core subsets from the original unlearning corpus.

\textbf{Distribution-based}.
Text-based objectives need to provide labels from limited vocabulary, which restricts optimization space. 
In order to achieve more fine-grained unlearning, some methods aim to make the model's output distribution converge to a reference distribution that aligns with the unlearning goals. 
 uses a uniform distribution .
Despite using existing distribution like uniform over the entire vocabulary~\cite{yuan2024closer}, 
reference distribution can establish either by modifying data or manipulating logits.
WHP~\cite{Eldan2023Whos} substitutes the unlearning target in original data with unrelated entities to get general knowledge distribution that does not contain sensitive information. 
WPU~\cite{Liu2024Rethinking} improves upon WHP by incorporating diverse substitute entities, performing entity name restoration, and augmenting input prompts. 
On the other hand, RKLD~\cite{Wang2025Balancing} taking the difference between the logits of a model finetuned on the unlearning set and that of original model as reference. 
Similar logits-aware approach is also adopted by Obliviate~\cite{russinovich2025obliviate} and PerMU~\cite{wang2025erasing}.
Distribution-based objectives generally shorten distance to target distribution via minimizing divergence. The most popular choice is KL divergence, but there are also methods using reverse KL~\cite{Wang2025Balancing}, JS divergence~\cite{Singh2025Unlearninga} or f-divergence~\cite{Wang2024LLM}.

\textbf{Activation-based}.
Both text-based and distribution-based methods treat the entire model as a black box, computing losses at the output level and perform back propagation, which is inefficient in many cases. 
% Moreover, these approaches may overlook that different parameters may contribute differently on generating knowledge to unlearn.
Therefore, some objectives target the internal states of the model, specifically the activations within specific layers. 
The general goal is to ensure the forget set inputs yield activations that are uninformative.
RMU~\cite{Li2024WMDP} combines a forget loss that perturbs hidden activations of harmful data towards a fixed random direction. 
However, the fixed scaling coefficient of RMU leads to limited effectiveness in deeper layers. 
To overcome this issue, ~\citet{huu2024effects} introduces an adaptive scaling coefficient proportional to the $l_2$-norm of the original representation. 
Similarly, LUNAR~\cite{shen2025lunar} aims to redirect the forget data’s activations into a refusal region, so that the model consistently produces safe refusal responses.
\citet{Guo2024Robust} and \citet{Wang2024Large} take advantage of mechanism interpretability by constructing the activation of the expected answer after unlearning, and reversely calculating the closed-form solution of parameter updates.

% \textbf{Parameter-based}:
% Instead of using a loss function to fine-tune, these methods directly manipulate the model's parameter space. 
% Some directly project parameters into a subspace where the information from the forget set is theoretically eliminated.
% For instance, \textit{UNLEARN}~\cite{Lizzo2024UNLEARN} conceptualizes knowledge as a weight-space subspace.
% It first learns low-rank matrices to represent the forget information, then applies a Gram-Schmidt-like orthogonalization against retained-task subspaces, and finally subtracts the resulting forget-subspace from the model.
% Similarly, \textit{Ethos}~\cite{Gao2024Ethos} employs SVD to construct an orthogonal basis, projects a task vector obtained from fine-tuning on undesired data, and isolates components with large singular values as representations of harmful knowledge.
% Then a new, purified task vector is reconstructed and subtracted from the model.
% Another class of methods is inspired by the success of Task Vectors in knowledge editing~\cite{Ilharco2022Editing}, which first fine-tuning an intermediate model and then arithmetically combining its parameters with those of the original model. 
% SKU~\cite{Liu2024Safer} fine-tunes a "bad model" to obtain a parameter offset that is antithetical to the unlearning goal. 
% This offset is then subtracted from the original model's parameters to yield a harmless unlearned model. 

\textbf{Multi-Objective Combination}.
Beyond unlearning objective itself, most methods add an loss term of retain set in practice to avoid degradation of general performance~\cite{Maini2024TOFU, Yao2024Large, Wang2023KGA, yuan2024closer, Lu2024Eraser}. 
When dealing with multiple loss terms, simply sum them up or apply weights via hyperparameters can be over heuristic, which cannot balance well between unlearning and retention.
NGDiff~\cite{Bu2024Unlearning} treats the combination between objectives as a multi-task optimization problem, achieving a better trade-off through precise normalization and dynamic learning rates. MOLLM~\cite{Pan2025MultiObjective} computes a common descent direction in the dual gradient space, yielding an update that simultaneously reduces influence of the target knowledge while preserving overall utility.

% \begin{table}
% \centering
% \begin{tabular}{lll} 
% \toprule
% Based on                      & Update Goal                 & Methods              \\ 
% \midrule
% \multirow{2}{*}{Text}         & Reduce Forget set Prob      & GA, NPO, FPGA, LLMU  \\ 
% \cmidrule{2-3}
%                               & Improve Substitute set Prob & RGD, DPO, AltPO      \\ 
% \midrule
% \multirow{2}{*}{Distribution} & Existing Distribution       & ME                   \\ 
% \cmidrule{2-3}
%                               & Synthetic Distribution      & WPU, RKLD            \\ 
% \midrule
% \multirow{2}{*}{Activation}   & Existing Activation         & RMU, Adaptive RMU    \\ 
% \cmidrule{2-3}
%                               & Synthetic Activation        & LUNAR                \\ 
% \midrule
% \multirow{2}{*}{Parameter}    & Subspace Projection         & Ethos, UNLEARN       \\
% \cmidrule{2-3}
%                               & Vector Merging              & SKU, SSU             \\
% \bottomrule
% \end{tabular}
% \end{table}

\subsubsection{Reinforcement Learning (RL)}
\label{sec:ft_rl}
Reinforcement learning (RL) serves as a foundational methodology in the training of LLMs, enabling them to learn decision-making policies by optimizing the cumulative rewards obtained through environmental interactions. 
A critical element in applying RL to LLM unlearning involves the design of reward mechanisms tailored to the unlearning scenario.

In pioneering work, Quark~\citep{Lu2022QUARK} quantifies undesirable attributes by converting the continuous reward into discrete quantiles
%discretizing reward values into predefined intervals 
and incorporating them as additional reward tokens into the input prompt during RL training. 
Another approach, DeMem~\citep{kassem2023preserving}, employs the negative BERTScore~\citep{zhang2019bertscore} as a reward signal, which measures the dissimilarity between the model-generated outputs and ground-truth references.
Beyond similarity-based rewards, RULE~\citep{zhang2025rule} integrates refusal behaviors for data targeted for unlearning into the reward function. 
Additionally, RULE introduces a rejection steering mechanism for warm-starting the training process, along with a boundary set comprising high-quality hard negatives to sharpen the learning signals near the decision boundary.

As noted in several studies~\citep{chu2025sft, chen2025two}, RL exhibits a superior generalization capability to unseen data compared to SFT, extending its effectiveness beyond the specific queries encountered during training. 
A further advantage is that RL-based unlearning operates without altering the underlying loss function, enabling its integration into established RL training pipelines, such as Reinforcement Learning from Human Feedback (RLHF)~\citep{bai2022training, ouyang2022training} or RL for enhancing reasoning capabilities~\citep{guo2025deepseek}. 
However, a widely acknowledged limitation of RL is its relatively slow convergence rate and inherent training instability~\citep{chen2025two}.

\subsubsection{Localizing Parameters}
\label{sec:ft_local}
\begin{table}[t]
    \footnotesize
    \centering
    \begin{tabular*}{.99\linewidth}{@{\hspace{10pt}\extracolsep{\fill}} l l l l @{\hspace{10pt}}}
            \toprule
            \multicolumn{2}{c}{Based on} & Method & Description\\
            \toprule
            \multirow{9}[4]{*}{\makecell{Data\\based}} & \multirow{6}[1]{*}{Gradient} 
            & DEPN~\citep{Wu2023DEPN}, SURE~\citep{zhang2025catastrophic} & $\nabla\mathcal{L}_f$\\
            & & SSU~\citep{Dou2024Avoiding} & $+$ gradient of random labeling loss.\\
            & & \citet{Stoehr2024Localizing} & $+$ gradient of KL divergence of output on retain set before/after unlearning.\\
            & & MemFlex~\citep{Tian2024Forget} & $+$ cosine similarity of $\nabla\mathcal{L}_f$ and $\nabla\mathcal{L}_r$.\\
            & & WAGLE~\citep{jia2024wagle} & $+$ element-wise product of $\nabla\mathcal{L}_f$ and $\nabla\mathcal{L}_r$.\\
            & & KLUE~\citep{yang2025faithun} & $+$ superficial knowledge regularization.\\
            %& & PCGU~\citep{Yu2023Unlearning} & Cosine similarity of gradients of an input pair.\\
            %\cmidrule(lr){2-4}
            %& DEPN~\citep{Wu2023DEPN} & \multirow{2}{*}{Neuron-level} & Neuron privacy scores; zero-out top ones \\
            %& KLUE~\citep{yang2025faithun} & & Target neurons linked to knowledge attribution \\
            \cmidrule(lr){2-4}
            & \multirow{3}{*}{Activation} & Selective Pruning~\cite{pochinkov2024dissecting} & four statistics of activations when processing forget versus retain data.\\
            & & REVS~\citep{Ashuach2024REVS} & Combination of activation strength and token association.\\
            & & FALCON~\citep{hu2025falcon} & Mutual information of activations of the unlearn and retain set.\\
            \toprule
            \multirow{4}[2]{*}{\makecell{Data\\free}} & \multirow{2}{*}{Heuristics} & RMU~\citep{Li2024WMDP} & \multirow{2}{*}{Experimental observation and hyperparameter search optimization.} \\
            & & Adaptive RMU~\citep{huu2024effects} & \\
            \cmidrule(lr){2-4}
            & \multirow{2}{*}{Mechanism} & LUNAR~\cite{shen2025lunar} & \multirow{2}{*}{Knowledge storage mechanism (down-projection matrix of MLP layers)~\citep{meng2022locating}.}\\
            & & LaW~\citep{Wang2024Large} &  \\
            %\cmidrule(lr){2-4}
            %& Interpretability & \makecell[l]{Mechanistic\\Unlearning~\citep{guo2024mechanistic}} & Causal tracing~\citep{meng2022locating}, attribution patching~\citep{nanda2023attribution}, probe~\citep{geva2023dissecting}, path patching~\citep{goldowsky2023localizing}.\\
            % \cmidrule(lr){2-5}
            % & \multirow{2}{*}{Subspace}
            % & UNLEARN~\citep{Lizzo2024UNLEARN} & \multirow{2}{*}{Subspace} & Subtract forget subspace via low-rank projection \\
            % & & Ethos~\citep{Gao2024Ethos} & & SVD-based purification of task vectors \\
            \bottomrule
        \end{tabular*}
    \caption{Outline of different parameter selecting methods. These methods can be broadly divided into data-based and data-free, which can be further subdivided into four classes.}
    \label{tab:local}
\end{table}

Simply fine-tuning of all parameters in a model often leads to issues such as high computational costs and potential performance degradation~\citep{Dou2024Avoiding}.
In several studies on interpretability and model editing~\citep{meng2022locating,Geva2021Transformer}, researchers have demonstrated that knowledge is associated with specific model weights, thus proposing methods to locate relevant parameters for more efficient updates.
Various techniques, including causal tracing~\citep{meng2022locating}, attribution patching~\citep{nanda2023attribution}, probing~\citep{geva2023dissecting}, and path patching~\citep{goldowsky2023localizing}, have been directly applied in research on unlearning~\citep{guo2024mechanistic}.
Furthermore, depending on the specific unlearning scenarios and objectives, numerous studies have proposed different strategies for parameter localization.
Based on whether task-specific data are required, we categorize these methods into two distinct classes, as summarized in Table \ref{tab:local}.

\paragraph{Data-based.}
There are various ways to select certain layers or neurons within the model. 
The most common selection criterion is the \textbf{loss gradient} w.r.t. the model parameters, calculated on the unlearn set ($D_u$), such as approaches in DEPN~\citep{Wu2023DEPN} and SURE~\citep{zhang2025catastrophic}.
It is based on the intuition that parameters with larger gradient magnitudes are more influential and should be prioritized for updates.
%DEPN~\citep{Wu2023DEPN} calculates the gradient of the loss function of the unlearn set and selects the top-k neurons.
As an updated work, SSU~\citep{Dou2024Avoiding} adds a random labeling loss to define a composite loss function, which is a commonly used data augmentation to enhance the stability~\citep{neelakantan2015adding}.
Meanwhile, several works consider the retain set when selecting parameters, reducing the impact of parameter updates on the retain data~\citep{Stoehr2024Localizing,Stoehr2024Localizing,Tian2024Forget} (refer to Table~\ref{tab:local} for details).
% They try to minimize terms including gradient of the KL divergence of output on retain set before and after unlearning~\citep{Stoehr2024Localizing}, cosine similarity of gradients of unlearn loss and retain loss~\citep{Tian2024Forget}, element-wise product of gradients of unlearn loss and retain loss~\citep{jia2024wagle}.
Furthermore, \citet{yang2025faithun} point out that different questions may share the same answer and should avoid unconditionally unlearning the answer regardless of the context.
Thus, they propose KLUE, which introduces a superficial knowledge regularization for accurate parameter localization.
%However, methods differ significantly in how they compute and utilize these gradient signals, showing a clear trajectory from using simple magnitudes to more complex relational criteria, and a shift in granularity from individual weights to entire neurons.

An alternative to gradient-based methods is to directly analyze the \textbf{activation} of the model's intermediate layers, which provides a direct lens into the model's internal knowledge representation, bypassing the computation need for backpropagation.
%Although these methods share a common focus, they differ significantly in how they interpret these activations to select parameters for modification.
The method of selective pruning~\cite{pochinkov2024dissecting} calculates an importance score for each neuron based on four statistics of its activations when processing unlearn versus retain data. 
In addition to the activation strength, REVS~\citep{Ashuach2024REVS} also considers the rank of a target token when projecting the neuron to the vocabulary space by unembedding matrix, where a lower rank value indicates a stronger association between the target token and the neuron.
They show that the combination outperforms methods based solely on activations, token associations, and gradients.
%Neurons that are significantly more active on the forget set than on the retain set receive a high score and are iteratively pruned from the model.
%\citet{Ashuach2024REVS} directly examines the hidden states of the model's intermediate layers. 
%Their approach, known as REVS, projects these hidden states and corresponding model weights into the logit space~\cite{nostalgebraist2020interpreting}. 
%Layers and neurons where the target tokens from $\mathcal{D}_f$ rank highly are selected and then iteratively adjusted to desired lower ranks.
Another approach, FALCON~\cite{hu2025falcon}, uses mutual information of activations of the unlearn and retain set, to identify layers where the hidden representations of forget and retain knowledge are least entangled, targeting these specific layers for modification.
%\textcolor{red}{TBD here.}

\paragraph{Data-free.}
Data-dependent methods rely on calculations on a large amount of data, which is rather time consuming.
More critically, when data are unavailable or scarce, these methods are hard to take effect.
Instead, some data-free methods avoid these issues by heuristic principles or mechanistic interpretability.
\citet{Li2024WMDP} observe that it is sufficient to compute the loss only on layer $\ell$ and update gradients only on layers $\ell-2$, $\ell-1$ and $\ell$, and perform a hyperparameter search over the layer to select the best layer for fine-tuning.
This setting is followed by \citet{huu2024effects}.
Additionally, inspired by insights into knowledge storage mechanism of LLMs \citep{meng2022locating}, LUNAR~\citep{shen2025lunar} and LaW~\citep{Wang2024Large} select the down-projection matrix of the MLP layers to update.
% In general, heuristic approaches exploit simple but effective rules to determine where to intervene, while mechanistic approaches aim to trace and manipulate the specific internal circuits responsible for generating knowledge. 
In general, heuristic approaches rely on simple and effective rules to select intervention sites, whereas mechanistic approaches target the specific internal circuits responsible for knowledge generation.

\subsubsection{Incorporating New Structure}
\label{sec:ft_new}
\begin{figure}[t]
    \centering
    \scriptsize
    \begin{tikzpicture}
        %\node at (-4,0.6) {\small \em Add parameters};
        \node (layer4-1) at (-5,0) [rectangle, draw=blue, fill=lightblue, minimum width=40pt, minimum height=15pt, inner sep=3pt, rounded corners=5pt, line width=1pt] {Layer $\ell-1$};
        \node (layer4-2) at (-5,-1.5) [rectangle, draw=blue, fill=lightblue, minimum width=40pt, minimum height=15pt, inner sep=3pt, rounded corners=5pt, line width=1pt] {Layer $\ell$};
        \node (layer4-3) at (-4.7,-3) {\small (a) EUL \& GRUN};
        \draw [thick,->,line width=1pt] ([yshift=10pt]layer4-1.north) -- node[draw=none,fill=none] {} (layer4-1.north);
        %\draw [thick,->,line width=1pt] (layer4-1.south) -- node[draw=none,fill=none] {} (layer4-2.north);
        \draw [thick,->,line width=1pt] (layer4-2.south) -- (-5,-2.7);

        \node (unl) at (-4,-0.75) [rectangle, draw=red, fill=lightred, minimum width=30pt, minimum height=10pt, inner sep=3pt, rounded corners=5pt, line width=1pt] {Fine-tuning module};
        \draw [thick,->,line width=1pt] (layer4-1.east) -| node[draw=none,fill=none] {} (unl.north);
        \draw [thick,->,line width=1pt] (unl.south) |- node[draw=none,fill=none] {} (layer4-2.east);

        %%%%%%%%%%%%%%%%%%%%%%%%%%%
    
        \node (layer1-1) at (-1,0) [rectangle, draw=blue, fill=lightblue, minimum width=40pt, minimum height=15pt, inner sep=3pt, rounded corners=5pt, line width=1pt] {Layer $\ell-1$};
        \node (layer1-2) at (-1,-1.5) [rectangle, draw=blue, fill=lightblue, minimum width=40pt, minimum height=15pt, inner sep=3pt, rounded corners=5pt, line width=1pt] {Layer $\ell$};
        \node (plus) at (-1,-2.1) [circle, draw=black, fill=none, inner sep=0pt, line width=1pt] {\Large $+$};
        \node (layer1-3) at (-0.2,-3) {\small (b) LoRA-based};
        \draw [thick,->,line width=1pt] ([yshift=10pt]layer1-1.north) -- node[draw=none,fill=none] {} (layer1-1.north);
        \draw [thick,->,line width=1pt] (layer1-1.south) -- node[draw=none,fill=none] {} (layer1-2.north);
        \draw [thick,-,line width=1pt] (layer1-2.south) -- node[draw=none,fill=none] {} (plus.north);
        \draw [thick,->,line width=1pt] (plus.south) -- (-1,-2.7);

        \node (a) at (0.8,-1.3) [trapezium, draw=red, fill=lightred, minimum width=30pt, minimum height=10pt, trapezium left angle=120, trapezium right angle=120] {Matrix A}; 
        \node (b) at (0.8,-1.7) [trapezium, draw=red, fill=lightred, minimum width=30pt, minimum height=10pt, trapezium left angle=60, trapezium right angle=60] {Matrix B};
        \draw [thick,->,line width=1pt] ([yshift=-13pt]layer1-1.south) -| node[draw=none,fill=none] {} (a.north);
        \draw [thick,->,line width=1pt] (b.south) |- node[draw=none,fill=none] {} (plus.east);

        %%%%%%%%%%%%%%%%%%%%%%%%%%%

        \node (layer3-1) at (4.3,0) [rectangle, draw=blue, fill=lightblue, minimum width=40pt, minimum height=15pt, inner sep=3pt, rounded corners=5pt, line width=1pt] {Layer $\ell-1$};
        \node (router) at (4.3,-0.7) [rectangle, draw=red, fill=lightred, minimum width=30pt, minimum height=10pt, inner sep=3pt, rounded corners=5pt, line width=1pt] {Router};
        
        %\node at (5.8,-1.5) [rectangle, draw=blue, fill=lightblue, minimum width=40pt, minimum height=15pt, inner sep=3pt, rounded corners=5pt, line width=1pt] {Layer $\ell$};
        \node (layer3-2) at (5.4,-1.5) [rectangle, draw=red, fill=lightred, minimum width=25pt, minimum height=15pt, inner sep=3pt, rounded corners=5pt, line width=1pt] {Layer $\ell$};
        \node at (4.3,-1.5) [rectangle, draw=red, fill=lightred, minimum width=25pt, minimum height=15pt, inner sep=3pt, rounded corners=5pt, line width=1pt] {Layer $\ell$};
        %\node at (4.6,-1.5) [rectangle, draw=blue, fill=lightblue, minimum width=40pt, minimum height=15pt, inner sep=3pt, rounded corners=5pt, line width=1pt] {Layer $\ell$};
        \node at (3.2,-1.5) [rectangle, draw=red, fill=lightred, minimum width=25pt, minimum height=15pt, inner sep=3pt, rounded corners=5pt, line width=1pt] {Layer $\ell$};
        \node (layer3-3) at (4.3,-3) {\small (c) LOKA};
        \draw [thick,->,line width=1pt] ([yshift=10pt]layer3-1.north) -- node[draw=none,fill=none] {} (layer3-1.north);
        \draw [thick,->,line width=1pt] (layer3-1.south) -- node[draw=none,fill=none] {} (router.north);
        \draw [thick,->,line width=1pt] (router.east) -| node[draw=none,fill=none] {} (layer3-2.north);
        \draw [thick,->,line width=1pt] (layer3-2.south) --++(0,-0.3) -| (4.3,-2.7);

    \end{tikzpicture}
    \caption{Illustration of three different approaches of incorporating new structure. \textcolor{blue!90!black}{Blue part} denotes the frozen parameters and \textcolor{red!90!black}{red part} denotes the parameter available for fine-tuning.}
    \label{fig:extra-params}
\end{figure}
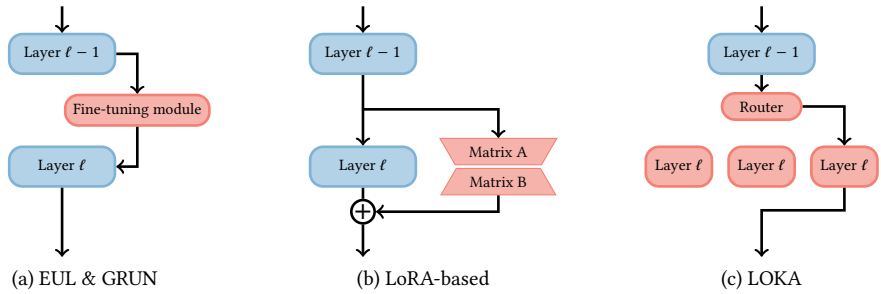

This type of method generally maintains the original ability of the model by freezing the existing parameters, and achieves forgetting by introducing new parameters or auxiliary structures.
A straightforward idea is to insert a new module between two layers of the model and only fine-tune this module, including EUL~\citep{Chen2023Unlearn} and GRUN~\citep{ren2025general}, which is illustrated in Figure \ref{fig:extra-params}(a).
This module has significantly fewer parameters compared to the original model, sometimes combined with structures like soft gate functions to improve performance~\citep{ren2025general}.
To deal with a sequence of unlearning requests, EUL and GRUN train a separate module on each unlearn task and design a fusion mechanism to merge all modules.
More research focuses on Low-Rank Adaptation (LoRA) \citep{hu2022lora}, which adds LoRA adapters to the model (Figure~\ref{fig:extra-params}(b)). 
%\citet{zhang2023composing} define negation operators for PEMs and applied it to the relevant PEM to achieve unlearning.
%Furthermore, \citet{hu2024separate} propose Extraction-before-Subtraction (Ext-Sub), which extracts the deficiency capability vector through the integration of ``expert'' PEM and ``anti-expert'' PEM, and subtracts only the deficiency vector.
%Although PEFT methods such as LoRA are efficient and widely adopted, 
However, standard LoRA lacks sufficient plasticity and often performs poorly in selective unlearning scenarios~\citep{cha2024robust}, which is followed by several key enhancements. 
\citet{cha2024robust} introduce Fisher-weighted Initialization of Low-rank Adapters (FILA). 
Meanwhile, \citet{gao2024large} address the challenge of continuous unlearning requests in practical settings. 
They employ an orthogonal regularization loss to disentangle different unlearning tasks within a single LoRA adapter and additionally train an out-of-distribution (OOD) detector to modulate the adapter activation based on the relevance of test samples to unlearned data.

In the updated work, LOKA~\citep{zhang2025resolving} introduces multiple storage modules to store distinct knowledge, effectively mitigating conflicts in LLM updating and improving storage efficiency. 
During training, input knowledge is allocated to the appropriate knowledge memories through similarity-aware knowledge mapping. 
During inference, a learning-based router dynamically activates the most relevant memory module according to the input prompt, enabling context-aware and conflict-minimized generation, which is illustrated in Figure \ref{fig:extra-params}(c).

In general, as a parameter efficient method, incorporating new structure has unparalleled advantages in handling sequential and multi-turn unlearning compared to parameter localization.
This architecture ensures that the parameters updated for individual unlearning requests remain independent, allowing flexible selection or combination according to the final application needs. 
More critically, through parameter integration methods such as fusion mechanisms or learnable routers, it alleviates two crucial problems in continual parameter fine-tuning: catastrophic forgetting of previous knowledge~\citep{french1999catastrophic} and knowledge interference between different rounds~\citep{zhang2025resolving}. 
However, this plug-in architecture presents several limitations. 
Firstly, its adaptability to downstream tasks may be constrained. 
Furthermore, since unlearning is confined solely to the integrated auxiliary structures, deactivating these components can effectively circumvent the defense mechanism, thereby allowing the recovery of unlearned content from the original model~\citep{shen2025lunar}.  

\subsubsection{Composite Approaches}
\label{sec:ft_others}
A small portion of post-training unlearning methods integrate multiple techniques discussed in the preceding subsections.
%do not update the parameters through learning. 
We review several notable approaches here as a complementary perspective, categorizing them into (1) parameter arithmetic operations and (2) SAE-based methods.

Several studies explore direct \textbf{arithmetic operations} on model parameters. 
Inspired by advances in task vectors for knowledge editing~\cite{Ilharco2022Editing}, some methods fine-tune an intermediate model and combine its parameters arithmetically with those of the original model. 
For instance, SKU~\cite{Liu2024Safer} fine-tunes a ``bad model'' to obtain a parameter deviation that opposes the unlearning objective. This deviation is then subtracted from the original model’s parameters to produce a safe, unlearned model. 
A similar strategy is adopted in~\citet{Eldan2023Whos}.

Since training a model of comparable size to the original is computationally intensive, two key refinements have been proposed. 
First, fine-tuning can be performed using parameter-efficient fine-tuning (PEFT) techniques, where unlearning is achieved by applying negation operations to relevant parameter-efficient modules (PEMs)~\citep{zhang2023composing}. 
To further mitigate the risk of degrading general model capabilities, \citet{hu2024separate} combine an ``expert'' PEM with an ``anti-expert'' PEM and derive a general capability vector for preservation.
Second, as an alternative to fine-tuning, approximate negative models can be derived via subspace decomposition and projection techniques, such as the Gram–Schmidt orthogonalization used in UNLEARN~\citep{Lizzo2024UNLEARN} and the singular value decomposition (SVD) applied in Ethos~\citep{Gao2024Ethos}.

Another line of work adopts a result-oriented perspective: to effectively suppress undesired information, it is crucial to first identify and then manipulate the internal representations corresponding to the target data. 
Several studies integrate \textbf{sparse autoencoders (SAEs)}~\citep{ng2011sparse} into specific model layers to enhance interpretability and isolate relevant features. 
For example, \citet{farrell2024applying} identify features that strongly activate on the unlearn set while minimally impacting the retain set, and then clamp their activations to negative values during inference. 
Similarly, \citet{wu2024codeunlearn} introduce a trainable codebook between the encoder and decoder of an SAE. 
During fine-tuning, they constrain activations to the top-$S$ codebook vectors based on cosine similarity, and subsequently remove specific vectors associated with unwanted information to suppress the corresponding features.

\subsection{Inference-Time Unlearning}

\begin{table}[t]
    \footnotesize
    \centering
    \begin{tabular*}{.99\linewidth}{@{\hspace{10pt}\extracolsep{\fill}} l l l l @{\hspace{10pt}}}
            \toprule
            \multicolumn{2}{c}{Modifying} & Method & Description\\
            \toprule
            \multirow{7}{*}{Input} & \multirow{5}{*}{Token} & Prompting method~\citep{thaker2024guardrail} & Involve crafting specific instructions or system prompts. \\
            & & ICUL~\citep{Pawelczyk2024Context} & Label flipping disrupts the original association. \\
            & & ICKU~\citep{takashiro2025answer} & Uses \texttt{<UNL>} and \texttt{</UNL>} to encapsulate target knowledge. \\
            & & RAG-based~\citep{wang2024when} & Modify the knowledge base of RAG to simulate unlearning. \\
            & & \citet{muresanu2024unlearnable} & retrieve representative examples through quantized k-means clustering.\\
            \cmidrule(lr){2-4}
            & \multirow{2}{*}{Embedding} & ECO~\citep{Liu2024Large} & Classify prompts \& selectively corrupts token embeddings.\\
            & & SPUL~\citep{Bhaila2024Soft} & Optimize soft prompt tokens to induce unlearning.\\
            \toprule
            \multirow{5}[3]{*}{Output} & \multirow{2}{*}{Token} & Filtering method~\citep{thaker2024guardrail} & Screen the initial output and remove unwanted information.\\
            & & ALU~\citep{sanyal2025agents} & Four agents collaborate sequentially to sanitize responses.\\
            \cmidrule(lr){2-4}
            & \multirow{3}{*}{Logit} & $\delta$-UNLEARNING~\citep{Huang2024Offset} & Compute logit offset between two small models.\\
            & & ULD~\cite{Ji2024Reversing} & Subtract logits from a model with reversed training objectives.\\
            & & DExperts~\citep{liu2021dexperts} & Use 2 expert model to recalculating token probability when decoding. \\
            \bottomrule
        \end{tabular*}
    \caption{Outline of different inference-time unlearning methods. These methods operate by altering input or output content at token or embedding/logit level during the inference phase.}
    \label{tab:infer-time}
\end{table}
\label{sec:infer-time}
% The methods introduced above require modification on parameters of the original model, requiring relatively large computing resources. 
% In contrast to modifying parameters, inference-time unlearning methods choose to modify input or output content during the inference phase, which results in low requirements for computing devices and wider usage scenarios.
In contrast to the aforementioned approaches, which necessitate modifications to the parameters of the original model and consequently demand substantial computational resources, inference-time unlearning methods operate by altering \textbf{input} or \textbf{output} content during the inference phase. 
By avoiding direct updates to the core parameters, this strategy offers significant advantages: efficient adaptation, strong generalization across model architectures, and mitigation of catastrophic forgetting.
%Circumventing direct updates to the core parameters brings this strategy several advantages, which enables efficient model adaptation, demonstrates strong generalization across different architectures and mitigates catastrophic forgetting.
%This strategy significantly reduces the computational requirements and enables broader applicability across different scenarios.
More precisely, modification can be made at \textbf{token} or \textbf{embedding/logit} levels.
%\textbf{(1) token level}, which is human-readable, offers better interpretability, and is feasible for black-box models;
%\textbf{(2) embedding/logit level}, which is unreadable to humans, but typically enables more effective and efficient unlearning.
Refer to Table \ref{tab:infer-time} for a brief summary of all inference-time unlearning methods.

\paragraph{Input-based Methods}
This category modifies the input presented to the model to induce unlearning. 
An approach leverages in-context learning by inserting human-readable instructions or examples into prompts (\textbf{token level}), eliminating the need for parameter updates.
For \textbf{inserting instructions}, \citet{thaker2024guardrail} propose using system prompts that explicitly instruct the model to refuse to generating target content\footnote{For example, respond with ``I cannot provide information about [topic].''}.
To enhance efficiency, they apply a filter to detect input related to the target, activating the refusal prompt only when necessary.
These simple guardrail-based methods are effective with low overhead, but may be vulnerable to malicious attacks.

For \textbf{inserting examples}, \citet{Pawelczyk2024Context} propose In-Context Unlearning (ICUL), which constructs customized prompts with several input-label pairs, where an input in the unlearn set is flipped labeled and other inputs are correctly labeled.
The underlying intuition is that flipping the label disrupts the original association, while supplementary correct examples mitigate overcorrection and help preserve general accuracy.
To address hallucination issues in ICUL, \citet{takashiro2025answer} introduce In-Context Knowledge Unlearning (ICKU), which wraps target knowledge between special tokens \texttt{<UNL>} and \texttt{</UNL>}, enabling flexible unlearning during inference. 
Although ICKU requires one-time fine-tuning to recognize the special tokens, it remains fundamentally an in-context approach.

In addition to unlearning through in-context methods, knowledge can also be stored outside the model, and reasonable strategies can be adopted to \textbf{provide the correct samples} during each in-context learning.
%Another line of work integrates retrieval-augmented generation (RAG). 
\citet{wang2024when} propose a RAG-based framework where the model answers queries based on an external knowledge base. 
Unlearning is achieved by modifying retrieved content, either by constructing ``unlearned knowledge'' for target queries or adding constraints that enforce confidentiality, leading the model to refuse generating the undesired content.
\citet{muresanu2024unlearnable} investigate a sample selection mechanism that constructs prompts by retrieving representative examples from the training set. 
Their approach employs quantized k-means clustering to partition the data and retrieves samples nearest to each cluster centroid. 
The authors prove that, with high probability, removing a single data point does not perturb the resulting cluster structure, thereby enabling unlearning without requiring additional retraining or modification.

Token-level modification offers several advantages, such as human readability, superior interpretability, and compatibility with black-box models. 
In contrast, other approaches adjust inputs at the \textbf{embedding level} to achieve more effective and efficient unlearning. 
Similar to post-training techniques, these methods are optimization-based but shift the optimization objective from model parameters to the input embeddings. 
For example, ECO~\citep{Liu2024Large} employs zeroth-order optimization at the embedding level and improves efficiency by using a classifier to select only relevant tokens for adjustment. 
To avoid per-sample optimization, SPUL~\citep{Bhaila2024Soft} optimizes a small set of soft prompt tokens via a multi-objective loss function, which are then selectively appended to input queries to induce unlearning.

\paragraph{Output-based Methods}
While input-based methods are still susceptible to the inherent unpredictability and lack of controllability within the LLM, modifying the model's output provides a more direct and precise alternative.
%While input-based methods still face uncontrollable issues within the LLM, modifying the model's output offers as a more precise strategy.
%This category involves \textbf{modifying the model's output}.
%, either by filtering the initial output or by leveraging logit revision to achieve unlearning.
At the \textbf{token level}, a straightforward idea is filtering, where the initial output of the model are automatically screened and censored to remove unwanted information before being presented to users~\citep{thaker2024guardrail}. 
Moving beyond simple filtering, \citet{sanyal2025agents} propose ALU, which employs four specialized agents (Vanilla, AuditErase, Critic, and Composer) that collaborate sequentially to sanitize responses dynamically during inference. 
%This method achieves high unlearning success and scalability.

Similarly to the idea of modifying input at the embedding level, it is also possible to modify output at the \textbf{logit level}.
%For methods \textbf{modifying logits}, 
An idea is to intervene in the decoding process.
\citet{liu2021dexperts} propose DExperts, which combines a language model with ``expert'' and ``anti-expert'' models, recalculating token probability distributions at each decoding step to avoid generating unwanted content.
%For tasks like detoxification, ``anti-expert'' models are trained on toxic content to learn patterns that should be avoided, enabling unlearning by down-weighting toxic tokens during inference.
Another line of research achieves unlearning by leveraging the logit differences between the target model and the auxiliary models. 
\citet{Huang2024Offset} introduce $\delta$-UNLEARNING, which computes a logit offset using two small white-box models, one retained and one unlearned (via methods like gradient ascent or KL minimization). 
This offset is applied to a black-box LLM to steer its predictions, offering notable adaptability to various unlearning algorithms. 
While $\delta$-UNLEARNING requires training both retain and unlearn models, \citet{Ji2024Reversing} simplify it with the Unlearning from Logit Difference (ULD) method. 
ULD trains a single assistant model with reversed objectives to remember the unlearn set and forget the retain set, then subtracts its logits from the original model's outputs to induce unlearning. 
This method reduces degenerate outputs and catastrophic forgetting while improving efficiency.
%through techniques like LoRA. 
%\input{sec/4_benchmarks_and_datasets}
% !TeX root = ../main_csur.tex
\section{Evaluations}
\label{sec:eval}
% !TeX root = ../main_csur.tex
\begin{figure}[t]
    \centering
    \footnotesize
    \begin{tikzpicture}[align=left]
        \node (data) at (0,-0.5) [circle,draw=black,line width=1pt,rounded corners=5pt, minimum height=15pt, minimum width=40pt] {};
        \node at (0,-0.5){\includegraphics[width=1cm]{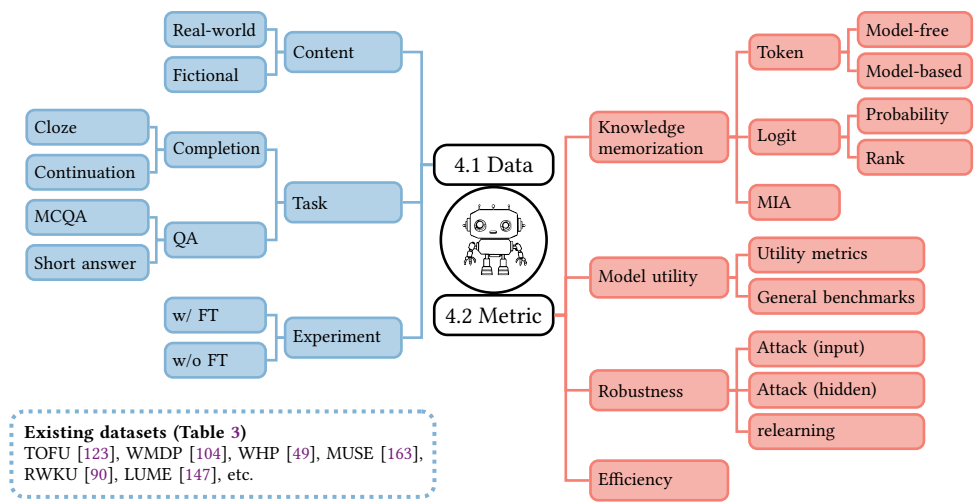}};
        \node (data) at (0,0.5) [rectangle,draw=black,line width=1pt,rounded corners=5pt, minimum height=15pt, minimum width=45pt] {\normalsize 4.1 Data};
        \node (metric) at (0,-1.5) [rectangle,draw=black,line width=1pt,rounded corners=5pt, minimum height=15pt, minimum width=45pt] {\normalsize 4.2 Metric};
        
        \node (content) at (-2,2) [rectangle, draw=blue, fill=lightblue, text width=38pt, inner sep=3pt, minimum height=15pt, align=left, rounded corners=3pt, line width=1pt] {Content};
        \node (task) at (-2,0) [rectangle, draw=blue, fill=lightblue, text width=38pt, inner sep=3pt, minimum height=15pt, align=left, rounded corners=3pt, line width=1pt] {Task};
        \node (exp) at (-2,-1.8) [rectangle, draw=blue, fill=lightblue, text width=38pt, inner sep=3pt, minimum height=15pt, align=left, rounded corners=3pt, line width=1pt] {Experiment};
        \draw [-,thick,draw=blue,line width=1pt] (data.west) --++ (-0.15,0) |- (content.east);
        \draw [-,thick,draw=blue,line width=1pt] (data.west) --++ (-0.15,0) |- (task.east);
        \draw [-,thick,draw=blue,line width=1pt] (data.west) --++ (-0.15,0) |- (exp.east);

        \node (real) at (-3.7,2.3) [rectangle, draw=blue, fill=lightblue, text width=32pt, inner sep=3pt, minimum height=13pt, align=left, rounded corners=3pt, line width=1pt] {Real-world};
        \node (fic) at (-3.7,1.7) [rectangle, draw=blue, fill=lightblue, text width=32pt, inner sep=3pt, minimum height=13pt, align=left, rounded corners=3pt, line width=1pt] {Fictional};
        \draw [-,thick,draw=blue,line width=1pt] (content.west) --++ (-0.1,0) |- (real.east);
        \draw [-,thick,draw=blue,line width=1pt] (content.west) --++ (-0.1,0) |- (fic.east);

        \node (comp) at (-3.7,0.7) [rectangle, draw=blue, fill=lightblue, text width=32pt, inner sep=3pt, minimum height=13pt, align=left, rounded corners=3pt, line width=1pt] {Completion};
        \draw [-,thick,draw=blue,line width=1pt] (task.west) -- ++(-0.1,0) |- (comp.east);
        \node (qa) at (-3.7,-0.5) [rectangle, draw=blue, fill=lightblue, text width=32pt, inner sep=3pt, minimum height=13pt, align=left, rounded corners=3pt, line width=1pt] {QA};
        \draw [-,thick,draw=blue,line width=1pt] (task.west) -- ++(-0.1,0) |- (qa.east);
        %\draw [-,thick,draw=blue,line width=1pt] (task.west) -| (-3,-0.6) -- node[above]{\tiny\textcolor{blue!80!black}{QA}} (-4.3,-0.6);
        \node (cloze) at (-5.4,1) [rectangle, draw=blue, fill=lightblue, text width=40pt, inner sep=3pt, minimum height=13pt, align=left, rounded corners=3pt, line width=1pt] {Cloze};
        \node (conti) at (-5.4,0.4) [rectangle, draw=blue, fill=lightblue, text width=40pt, inner sep=3pt, minimum height=13pt, align=left, rounded corners=3pt, line width=1pt] {Continuation};
        \draw [-,thick,draw=blue,line width=1pt] (comp.west) --++(-0.1,0) |- (conti.east);
        \draw [-,thick,draw=blue,line width=1pt] (comp.west) --++(-0.1,0) |- (cloze.east);
        
        \node (mcqa) at (-5.4,-0.2) [rectangle, draw=blue, fill=lightblue, text width=40pt, inner sep=3pt, minimum height=13pt, align=left, rounded corners=3pt, line width=1pt] {MCQA};
        \node (saqa) at (-5.4,-0.8) [rectangle, draw=blue, fill=lightblue, text width=40pt, inner sep=3pt, minimum height=13pt, align=left, rounded corners=3pt, line width=1pt] {Short answer};
        \draw [-,thick,draw=blue,line width=1pt] (qa.west) --++(-0.1,0) |- (mcqa.east);
        \draw [-,thick,draw=blue,line width=1pt] (qa.west) --++(-0.1,0) |- (saqa.east);

        \node (ft) at (-3.7,-1.5) [rectangle, draw=blue, fill=lightblue, text width=32pt, inner sep=3pt, minimum height=13pt, align=left, rounded corners=3pt, line width=1pt] {w/ FT};
        \node (no_ft) at (-3.7,-2.1) [rectangle, draw=blue, fill=lightblue, text width=32pt, inner sep=3pt, minimum height=13pt, align=left, rounded corners=3pt, line width=1pt] {w/o FT};
        \draw [-,thick,draw=blue,line width=1pt] (exp.west) --++ (-0.1,0) |- (ft.east);
        \draw [-,thick,draw=blue,line width=1pt] (exp.west) --++ (-0.1,0) |- (no_ft.east);

        \node (exist) at (-3.6,-3.35) [rectangle, dotted, draw=blue, fill=none, text width=150pt, inner sep=5pt, rounded corners=5pt, line width=1.5pt, align=left] {
            \textbf{Existing datasets (Table~\ref{tab:D&B})}\\
            TOFU~\citep{Maini2024TOFU}, WMDP~\citep{Li2024WMDP}, WHP~\citep{Eldan2023Whos}, MUSE~\citep{Shi2024MUSE}, RWKU~\citep{Jin2024RWKU}, LUME~\citep{ramakrishna2025lume}, etc.
        };

        %%%%%%%%%%%%%%%%%%%%%%%%

        \node (mem) at (2.2,0.87) [rectangle, draw=red, fill=lightred, text width=46pt, inner sep=3pt, minimum height=15pt, align=left, rounded corners=3pt, line width=1pt] {Knowledge memorization};
        %\node (util) at (2.4,-0.7) [rectangle, draw=red, fill=lightred, minimum width=48pt, minimum height=15pt, rounded corners=3pt, line width=1pt,anchor=west,align=left] {\makecell[l]{Model\\utility}};
        \node (util) at (2.2,-1) [rectangle, draw=red, fill=lightred, text width=46pt, inner sep=3pt, minimum height=15pt, align=left, rounded corners=3pt, line width=1pt] {Model utility};
        \node (rob) at (2.2,-2.5) [rectangle, draw=red, fill=lightred, text width=46pt, inner sep=3pt, minimum height=15pt, align=left, rounded corners=3pt, line width=1pt] {Robustness};
        \node (prac) at (2.2,-3.7) [rectangle, draw=red, fill=lightred, text width=46pt, inner sep=3pt, minimum height=15pt, align=left, rounded corners=3pt, line width=1pt] {Efficiency};
        %\draw [-,thick,draw=red,line width=1pt] (metric.east) -- node[above]{\tiny\textcolor{red!90!black}{Result}} ++(0.6,0) |- (mem.west);
        \draw [-,thick,draw=red,line width=1pt] (metric.east) --++ (0.15,0) |- (mem.west);
        \draw [-,thick,draw=red,line width=1pt] (metric.east) --++ (0.15,0) |- (util.west);
        \draw [-,thick,draw=red,line width=1pt] (metric.east) --++ (0.15,0) |- (rob.west);
        \draw [-,thick,draw=red,line width=1pt] (metric.east) --++ (0.15,0) |- (prac.west);
        %\draw [-,thick,draw=red,line width=1pt] (metric.south) |- node[yshift=30pt,right]{\tiny\textcolor{red!90!black}{Process}} (prac.east);

        %\draw [-,thick,draw=red,line width=1pt] (mem.east) -| (3.5,2) -- node[above]{\tiny\textcolor{red!80!black}{Output}} (4.3,2);
        %\draw [-,thick,draw=red,line width=1pt] (mem.east) -| (3.5,0.8) -- node[above]{\tiny\textcolor{red!80!black}{Logit}} (4.3,0.8);
        \node (token) at (4,2) [rectangle, draw=red, fill=lightred, text width=28pt, inner sep=3pt, minimum height=13pt, align=left, rounded corners=3pt, line width=1pt] {Token};
        \draw [-,thick,draw=red,line width=1pt] (mem.east) --++(0.1,0) |- (token.west);
        \node (no_model) at (5.6,2.3) [rectangle, draw=red, fill=lightred, text width=37pt, inner sep=3pt, minimum height=13pt, align=left, rounded corners=3pt, line width=1pt] {Model-free};
        \node (model) at (5.6,1.75) [rectangle, draw=red, fill=lightred, text width=37pt, inner sep=3pt, minimum height=13pt, align=left, rounded corners=3pt, line width=1pt] {Model-based};
        \draw [-,thick,draw=red,line width=1pt] (token.east) --++(0.1,0) |- (no_model.west);
        \draw [-,thick,draw=red,line width=1pt] (token.east) --++(0.1,0) |- (model.west);

        \node (logit) at (4,0.87) [rectangle, draw=red, fill=lightred, text width=28pt, inner sep=3pt, minimum height=13pt, align=left, rounded corners=3pt, line width=1pt] {Logit};
        \draw [-,thick,draw=red,line width=1pt] (mem.east) --++(0.1,0) |- (logit.west);
        \node (prob) at (5.6,1.15) [rectangle, draw=red, fill=lightred, text width=37pt, inner sep=3pt, minimum height=13pt, align=left, rounded corners=3pt, line width=1pt] {Probability};
        \node (rank) at (5.6,0.6) [rectangle, draw=red, fill=lightred, text width=37pt, inner sep=3pt, minimum height=13pt, align=left, rounded corners=3pt, line width=1pt] {Rank};
        \draw [-,thick,draw=red,line width=1pt] (logit.east) --++(0.1,0) |- (prob.west);
        \draw [-,thick,draw=red,line width=1pt] (logit.east) --++(0.1,0) |- (rank.west);

        \node (mia) at (4,0) [rectangle, draw=red, fill=lightred, text width=28pt, inner sep=3pt, minimum height=13pt, align=left, rounded corners=3pt, line width=1pt] {MIA};
        \draw [-,thick,draw=red,line width=1pt] (mem.east) --++(0.1,0) |- (mia.west);
        
        \node (umetric) at (4.56,-0.7) [rectangle, draw=red, fill=lightred, text width=60pt, inner sep=3pt, minimum height=13pt, align=left, rounded corners=3pt, line width=1pt] {Utility metrics};
        \node (ubench) at (4.56,-1.25) [rectangle, draw=red, fill=lightred, text width=60pt, inner sep=3pt, minimum height=13pt, align=left, rounded corners=3pt, line width=1pt] {General benchmarks};
        \draw [-,thick,draw=red,line width=1pt] (util.east) --++(0.1,0) |- (umetric.west);
        \draw [-,thick,draw=red,line width=1pt] (util.east) --++(0.1,0) |- (ubench.west);

        \node (ainput) at (4.56,-1.95) [rectangle, draw=red, fill=lightred, text width=60pt, inner sep=3pt, minimum height=13pt, align=left, rounded corners=3pt, line width=1pt] {Attack (input)};
        \node (ahidden) at (4.56,-2.5) [rectangle, draw=red, fill=lightred, text width=60pt, inner sep=3pt, minimum height=13pt, align=left, rounded corners=3pt, line width=1pt] {Attack (hidden)};
        %\node (mia) at (4.43,-2.5) [rectangle, draw=red, fill=lightred, text width=52pt, inner sep=3pt, minimum height=13pt, align=left, rounded corners=3pt, line width=1pt] {MIA};
        \node (relearn) at (4.56,-3.05) [rectangle, draw=red, fill=lightred, text width=60pt, inner sep=3pt, minimum height=13pt, align=left, rounded corners=3pt, line width=1pt] {relearning};
        \draw [-,thick,draw=red,line width=1pt] (rob.east) --++(0.1,0) |- (ainput.west);
        \draw [-,thick,draw=red,line width=1pt] (rob.east) --++(0.1,0) |- (ahidden.west);
        \draw [-,thick,draw=red,line width=1pt] (rob.east) --++(0.1,0) |- (relearn.west);
        %\draw [-,thick,draw=red,line width=1pt] (rob.east) --++(0.1,0) |- (mia.west);
    \end{tikzpicture}
    \caption{Evaluation Framework. It involves two parts: (1) data and (2) metrics. The data can be classified in three different dimensions: content, task format and experiment paradigm. Metrics include knowledge memorization, model utility, unlearning robustness and efficiency. Additionally, we include existing datasets with their features in Table \ref{tab:D&B}.}
    \label{fig:eval_frame}
\end{figure}

Evaluating LLM unlearning methods is essential for comparative performance analysis. 
% As illustrated in the first row of Figure \ref{fig:eval_frame}, a typical unlearning evaluation process consists of several stages. 
% First, a target model is selected and a subset of its knowledge is designated as the unlearning set. 
% This knowledge may be intrinsic to the pre-trained model or acquired through fine-tuning. 
% An unlearning algorithm is then applied to produce the unlearned model, which is subsequently compared against an expected reference model to assess the outcome. 
% Note that this expected model does not actually exist, but rather the ideal result we expect to achieve after unlearning.
This procedure raises two fundamental questions:
\begin{quote}
    \em
    (Q1) In which datasets are the experiments conducted?  \\
    (Q2) What metrics are used to quantify the results?  
\end{quote}
To address (Q1), Section \ref{sec:eval_data} examines the data from three dimensions, including task format, content, and experiment paradigm, along with commonly used benchmarks. 
To aid in benchmark selection, Table~\ref{tab:D&B} summarizes key features to offer an overview of existing benchmarks.
For (Q2), Section \ref{sec:eval_metric} categorizes the evaluation metrics into four classes based on the aspect of model behavior they assess: knowledge memorization, model utility, unlearning efficiency, and unlearning robustness.
Refer to Figure~\ref{fig:eval_frame} for an overview of this section.

\subsection{Data}
\label{sec:eval_data}
In Section~\ref{sec:def}, we introduce the definition of the unlearn set $\mathcal{D}_u$ and the retain set $\mathcal{D}_r$.
%the objective is to remove specific data, referred to as the unlearn set (shadow squares in Figure \ref{fig:eval_frame}), while preserving all other knowledge, collectively termed the \textbf{retain set} (white squares). 
In LLM unlearning, the retain set can be further categorized into the \textit{neighbor set} and \textit{world set} based on relevance to the unlearn set. 

The \textbf{neighbor set} consists of data that is semantically related yet distinct from the unlearn set. Common construction strategies include withholding a subset (e.g., 1\%, 5\%, or 10\% under the TOFU framework~\citep{Maini2024TOFU}) from unlearning, or manually curating content from related domains. 
For example,~\citet{Lynch2024Eight} extract mythology and film production details using GPT-4 when unlearning Harry Potter material, while~\citet{Shi2024MUSE} source related content from the Harry Potter FanWiki. 
As highlighted by~\citet{choi2025opt-out}, neighbor samples act as “hard positives,” helping the model discriminate between unlearn and retain knowledge. 
Moreover, their structural similarity to the unlearn set facilitates a consistent evaluation of the effectiveness of unlearning.

\textbf{World set} denotes the broad, general information acquired during pretraining, which is largely independent of the unlearn set. 
It is typically drawn from large-scale repositories such as Wikidata~\citep{vrandecic2014wikidata}, OpenWebText~\citep{peterson2019open} and FineWeb~\citep{penedo2024fineweb}. 
Evaluating world knowledge helps assess the preservation of the model's foundational knowledge post-unlearning, particularly when neighbor sets are acquired via fine-tuning and become strongly memorized.
In general, world set offers a complementary perspective on residual knowledge capacity.

It is worth noting that directly synthesizing or constructing these datasets may introduce several issues, such as information overlap between unlearn and retain sets~\citep{jiang2025holistic}, incomplete memorization of the unlearn set by the original model~\citep{ma2025unveiling}, and increased unlearning difficulty for data associated with minority groups~\citep{wei2024underestimated}. 
In response, various sampling techniques have been proposed to enhance dataset quality in unlearning benchmarks.

After understanding the composition of the dataset, we classify the data from three different perspectives and summarize the advantages of each feature in the Table~\ref{tab:D&B}(a).

\subsubsection{Task Format}
\begin{figure}
    \centering
    \begin{tikzpicture}
        \node (completion) at (-3.6,-0.1) [rectangle, draw=red, fill=none, minimum width=6.1cm, minimum height=4.3cm, inner sep=5pt, rounded corners=5pt, line width=1pt] {};
        \node at (-3.6,2.05) [fill=white,inner sep=2pt] {\textcolor{red}{\Large \em Text Completion}};
        \node (cloze) at (-3.6,1) [rectangle, draw=red, fill=lightred!50, inner sep=5pt, rounded corners=5pt, line width=1pt, text width=5.5cm, align=left] {
            {\large \textbf{Cloze}}\\
            \footnotesize \textbf{Question}: In The Shawshank Redemption, Andy Dufresne is played by \_\_\_ Robbins. \\
            \textbf{Ground truth}: Tim~\citep{Jin2024RWKU}\\~\vspace{-\baselineskip}
        };
        \node (continuation) at (-3.6,-1) [rectangle, draw=orange, fill=lightorange!50, inner sep=5pt, rounded corners=5pt, line width=1pt, text width=5.5cm, align=left] {
            \large\textbf{Continuation}\\
            \footnotesize\textbf{Question}: In his third year, they were required to buy a particular textbook for Care of Magical Creatures, a book that was notorious for\\
            \textbf{Ground truth}: being one of the most difficult and complex classes at Hogwarts.~\citep{Eldan2023Whos}\\~\vspace{-\baselineskip}
        };
        
        \node (qa) at (3.6,-0.1) [rectangle, draw=blue, fill=none, minimum width=7.9cm, minimum height=4.3cm, inner sep=0pt, rounded corners=5pt, line width=1pt] {};
        \node at (3.6,2.05) [fill=white,inner sep=2pt] {\textcolor{blue}{\Large \em Question \& Answer}};
        \node (multichoice) at (3.6,1) [rectangle, draw=green, fill=lightgreen!50, inner sep=5pt, rounded corners=5pt, line width=1pt, text width=7.3cm, align=left] {
            {\large \textbf{Multiple Choice}}\\
            \footnotesize \textbf{Question}: What mutation in the poliovirus IRES often arises during cell culture adaptation? A. G480A B. A103G C. C472U D. A181V\\
            \textbf{Ground truth}: C. C472U~\citep{Li2024WMDP}\\~\vspace{-\baselineskip}
        };
        \node (shortanswer) at (3.6,-1) [rectangle, draw=blue, fill=lightblue!50, inner sep=5pt, rounded corners=5pt, line width=1pt, text width=7.3cm, align=left] {
            {\large \textbf{Short Answer}}\\
            \footnotesize \textbf{Question}: Who is this celebrated LGBTQ+ author from Santiago, Chile known for their true crime genre work?\\
            \textbf{Ground truth}: The author in question is Jaime Vasquez, an esteemed LGBTQ+ writer who hails from Santiago, Chile and specializes in the true crime genre.~\citep{Maini2024TOFU}\\~\vspace{-\baselineskip}
        };
    \end{tikzpicture}
    \caption{Examples of different tasks. Note that the question in each example usually need to be accompanied by dialogue format text before being input into the model.}
    \label{fig:qa_example}
\end{figure}
\label{sec:eval_task}
Based on the data, the model needs to complete specific tasks for evaluation.
%While some early works~\citep{Pawelczyk2024Context,Bhaila2024Soft,Chen2023Unlearn} employ classification tasks, this approach is less common in the contemporary usage scenario of LLMs. 
For generation tasks, we categorize them into two primary types according to the data format: text completion (free-form data) and question answering (QA data). 
As illustrated in Figure \ref{fig:qa_example}, these are subdivided into four distinct subcategories. 

\textbf{Text completion} directly provides partial data from the unlearn set to the model, requiring the model to fill in the blanks (\textbf{cloze}), or to continue to generate complete sentences (\textbf{continuation}). 
Additionally, for masked language models (MLMs) such as BERT~\citep{devlin2019bert} and RoBERTa~\citep{liu2019roberta}, this can also be achieved by predicting the masked word~\citep{Chen2023Unlearn}.
Examples of these tasks are shown on the left side of Figure \ref{fig:qa_example}.
Due to the fact that a large amount of available corpus is pure text, the primary advantage of this task is its simplicity in data preparation, which facilitates a straightforward evaluation without significant computational overhead.
%However, most LLMs are autoregressive models that can only generate forwards, resulting in relative limitations on text completion tasks.
Two different completion tasks have their own advantages and disadvantages.
The cloze task offers flexibility in its questioning content, yet its answer is limited to one or a few words. 
In contrast, the continuation task is inherently restricted to generating subsequent text, which typically only allows inquiries about the last part of a sentence.
Advantages of different tasks are summarized in Table \ref{tab:D&B}(a).

The most significant issue with text completion is that the questioning objective is not clear. 
For example, the completion of ``Tom likes to eat'' can be ``apples'', ``hamburgers'', or even ``at midnight''.
\textbf{Question \& Answer (QA)} can solve this problem.
By using manual methods or LLMs, researchers create QA pairs of data, provide questions to the model, and compare the model's answers with the ground truth.
Depending on the type of question, it can be divided into \textbf{multiple choices} and \textbf{short answers}.
Refer to the right side of Figure \ref{fig:qa_example} for examples.
Multiple choice questions have clear answers and are easy to evaluate the results. 
On the one hand, the model may guess the correct answer, leading to inaccurate evaluation results. 
On the other hand, this can also be used as a potential attack method~\citep{Lynch2024Eight}, as the model is required to choose an answer from the options provided and cannot deceive by fabricating irrelevant content.
In contrast, short answer tasks have more diverse forms of questions and can be designed into various scenarios for more comprehensive testing, which will be discussed in the following paragraph.
%On the other hand, for short answers, the same problem that model's answers may not be completely consistent with the ground truth is existed and metrics in the previous paragraph are also needed.

Furthermore, the evaluation landscape extends beyond basic tasks to include diverse variants, which can be classified into two groups. 
The first focuses on prompt manipulation, such as translation~\citep{choi2024cross,lu2025learn,Jin2024RWKU,Lynch2024Eight}, rephrasing, reverse query, and synonym substitution~\citep{choi2025opt-out,sanyal2025agents}. 
The second designs structured scenarios, such as analogy completion or odd-one-out tasks~\citep{joshi2024towards}.
Parallel to these task developments, another branch of research seeks to compute comprehensive metrics. 
To mitigate the reliance on point estimates, \citet{Scholten2024Probabilistic} propose a probabilistic framework, which is calculated by extensively sampling the model generation. 
Other studies aggregate performance in numerous tasks through designed average scores~\citep{Liu2024Learning,Shi2024MUSE} or Cognitive Diagnosis Models (CDMs)~\citep{lang2025beyond}.

\subsubsection{Content}
%The content of the unlearn set can be either real or fictional.
From the perspective of content, the unlearn set may originate from \textbf{real-world sources}, such as the Harry Potter series~\citep{Eldan2023Whos}, or be \textbf{fictionally constructed}, as exemplified by the TOFU benchmark~\citep{Maini2024TOFU}. 
Real-world data exhibit richer content and more coherent logical relationships, thus being practically useful~\citep{Jin2024RWKU}.
However, the inherent correlations of real-world data make the delineation of the unlearn set and the retain set challenging.
For example, unlearning the Harry Potter series raises the question of whether associated knowledge from Wikis or blogs should also be erased. 
To address this issue, several studies employ fictional data generated via templates or LLMs~\citep{Maini2024TOFU,wu2024evaluating}.
Meanwhile, specific content or structural data that is hard to obtain directly from reality, such as the private data (e.g., phone number, address)~\citep{ramakrishna2025lume} or relationship graphs~\citep{Qiu2024PISTOL}, can also be generated.

\subsubsection{Experiment Paradigm}
In the experiment, datasets can be broadly categorized into two classes based on whether fine-tuning is required. 
In the first category, the models perform unlearning \textbf{without fine-tuning} on the target dataset, which simplifies experimental setup. 
This includes the following scenarios.
(1) The unlearn set is compiled directly from the model’s pretraining data, such as subsets derived from the Pile~\citep{gao2020pile}.
(2) The data are manually verified to be present in the model, as in RWKU~\citep{Jin2024RWKU}, ConceptVectors~\citep{Hong2024Intrinsic}, and RETURN~\citep{Liu2024Learning}.
However, verifying the presence of facts in LLMs remains challenging and may affect reliability.
(3) For security purposes, the model is required to erase certain knowledge regardless of its original presence, exemplified by WMDP~\citep{Li2024WMDP} and UNCD~\citep{lang2025beyond}.
In the second category, models are \textbf{first fine-tuned} on the full dataset before a subset is unlearned. 
This is essential when datasets are fictionally synthetic, such as TOFU~\citep{Maini2024TOFU}, EDU-RELAT~\citep{wu2024evaluating}, and PISTOL~\citep{Qiu2024PISTOL}, to ensure that the model acquires the target knowledge. 
Even for real-world corpora, fine-tuning helps to guarantee that the original model possesses knowledge of the unlearn set.
\begin{table}[!p]
    \footnotesize
    \centering
    \begin{minipage}{.99\linewidth}
        \centering
        \begin{tabular*}{\linewidth}{@{\hspace{5pt}\extracolsep{\fill}} l l l l @{\hspace{5pt}}}
            \toprule
            \multicolumn{2}{c}{Class} & \multicolumn{2}{c}{Advantages}\\
            \toprule
            \multirow{4}[3]{*}{Task} & Cloze & \multirow{2}{*}{(Free-form) simple data preparation.} &  Flexible position of questioning content.\\
            & Continuation &  & Long answer length.\\
            \cmidrule(lr){2-4}
            & MCQA & \multirow{2}{*}{(QA) clear questioning objectives.} & Unique answer, easy for evaluation\\
            & Short answer &  & Various forms and scenarios. \\
            \midrule
            \multirow{2}[1]{*}{Content} & Real-world & \multicolumn{2}{l}{Rich content, coherent logical relationships, practically useful.}\\
            %\cmidrule(lr){2-4}
            & Fictional & \multicolumn{2}{l}{Easy to separate the unlearn/retain set, flexible to construct required content and format.} \\
            \midrule
            \multirow{2}[1]{*}{Experiment} & W/o fine-tuning & \multicolumn{2}{l}{Low computational cost and simple experiment.}\\
            %\cmidrule(lr){2-4}
            & W/ fine-tuning & \multicolumn{2}{l}{Ensure that the original model memorize the unlearn set, continuous learning-unlearning scenario.}\\
            \bottomrule
        \end{tabular*}
        \label{tab:comparison}
        \subcaption{Comparison of different tasks formats, data contents and experiment paradigms.}
    \end{minipage}
    \begin{minipage}{.99\linewidth}
        \centering
        \begin{tabular*}{\linewidth}{@{\hspace{5pt}\extracolsep{\fill}} l@{}c@{}c@{}c l l m{110pt} @{\hspace{5pt}}}
            \toprule
            Benchmark & Real & Fic. & FT & Data (\textcolor{red!90!black}{Free-form}/\textcolor{blue!90!black}{QA}) & Tasks & Used by\\
            \toprule
            TOFU~\citep{Maini2024TOFU} & & \checkmark & \checkmark & \textcolor{blue!90!black}{200$\times$20 QA} & SAQA & \citep{joshi2024towards,yuan2024closer,Scholten2024Probabilistic,wang2025erasing,shen2025lunar,zhang2025resolving,gao2024large,cha2024robust,ren2025general,thaker2024guardrail,Huang2024Offset,Liu2024Large,Ji2024Reversing,sanyal2025agents,takashiro2025answer,jia2024wagle,xu2025relearn,Gu2024MEOW,ma2025unveiling,wang2025effective,Mekala2024Alternate,Jia2024SOUL,Wang2024Rethinking,chen2025soft,Liu2024Revisiting,Bu2024Unlearning,Wang2024LLM,fan2025towards,fan2025simplicity,Wang2024RKLD,Zhang2024Negative} \\
            \cmidrule(lr){1-7}
            WMDP~\citep{Li2024WMDP} & \checkmark &  & \ding{55} & \textcolor{red!90!black}{Papers \& passages} & 3,668 MCQA & \citep{che2025model,doshi2024does,wang2025erasing,farrell2024applying,ren2025general,thaker2024guardrail,Liu2024Large,Bhaila2024Soft,sanyal2025agents,kolbeinsson2025composable,jia2024wagle,hu2025falcon,huu2024effects,lucki2024adversarial,fan2025simplicity,Kadhe2024Split,deeb2025unlearning} \\
            \cmidrule(lr){1-7}
            WHP~\citep{Eldan2023Whos} & \checkmark &  & \checkmark & \textcolor{red!90!black}{3.1M tokens} & 300 Conti + 30 Cloze & \citep{Lynch2024Eight,Scholten2024Probabilistic,wang2025erasing,thaker2024guardrail,Ji2024Reversing,sanyal2025agents,jia2024wagle,Jia2024SOUL,chen2025soft}\\
            \cmidrule(lr){1-7}
            MUSE~\citep{Shi2024MUSE} & \checkmark &  & \checkmark & \textcolor{red!90!black}{4.4M+6.5M tokens} & Conti + SAQA & \citep{jiang2025holistic,wang2025erasing,zhang2025catastrophic,Yuan2024Robust,russinovich2025obliviate,Bu2024Unlearning,Wang2024LLM,fan2025simplicity,dong2024undial,zhang2025rule} \\
            \cmidrule(lr){1-7}
            RWKU~\citep{Jin2024RWKU} & \checkmark &  & \ding{55} & \textcolor{red!90!black}{200 celebrities} & 3,268 cloze + 2,879 SAQA & \citep{takashiro2025answer,Yuan2024Robust,zhang2025rule} \\
            \cmidrule(lr){1-7}
            CoTaEval~\citep{Wei2024Evaluating} & \checkmark &  & \checkmark & \textcolor{red!90!black}{1K + 1K passages} & 1.5K Conti + 1K SAQA & \citep{russinovich2025obliviate} \\
            \cmidrule(lr){1-7}
            KnowUnDo~\citep{Tian2024Forget} & \checkmark & \checkmark & \checkmark & \textcolor{blue!90!black}{2,649 QA} & SAQA & \citep{xu2025relearn} \\
            \cmidrule(lr){1-7}
            PISTOL~\citep{Qiu2024PISTOL} &  & \checkmark & \checkmark & \textcolor{blue!90!black}{4 Graphs (50,95)} & 95 SAQA & \citep{shen2025lunar} \\
            \cmidrule(lr){1-7}
            WPU~\citep{Liu2024Revisiting} & \checkmark & & \ding{55} & \textcolor{red!90!black}{100 people's Wiki} & 2,795 SAQA & \citep{Sinha2024UnStar} \\
            \cmidrule(lr){1-7}
            LUME~\citep{ramakrishna2025lume} & \checkmark & \checkmark & \checkmark & \textcolor{red!90!black}{1,387 documents} & 4,394 (Conti + SAQA) & SEMEval-2025 Task 4 \\
            \cmidrule(lr){1-7}
            {\scriptsize ConceptVectors}~\citep{Hong2024Intrinsic} & \checkmark &  & \ding{55} & \textcolor{red!90!black}{285$\times$10 paragraph} & 285$\times$10 Conti + 285$\times$10 SAQA & - \\
            \cmidrule(lr){1-7}
            EDU-RELAT~\citep{wu2024evaluating} &  & \checkmark & \checkmark & \textcolor{blue!90!black}{700 QA} & 11$\times$5 SAQA & - \\
            \cmidrule(lr){1-7}
            ELUDe~\citep{choi2025opt-out} & \checkmark &  & \checkmark & \textcolor{blue!90!black}{15,651+90,954 QA} & MCQA + SAQA & - \\
            \cmidrule(lr){1-7}
            FaithUn~\citep{yang2025faithun} & \checkmark &  & \ding{55} & \textcolor{blue!90!black}{664 QA} & 8,377 MCQA & - \\
            \cmidrule(lr){1-7}
            LLM Surgery~\citep{veldanda2024llm} & \checkmark & \checkmark & \checkmark & \textcolor{red!90!black}{180K+1B tokens} & 24,800 MCQA & - \\
            \cmidrule(lr){1-7}
            Restor~\citep{rezaei2024restor} & \checkmark &  & \checkmark & \textcolor{red!90!black}{3,000 passages} & 1,051 SAQA & - \\
            \cmidrule(lr){1-7}
            RETURN~\citep{Liu2024Learning} & \checkmark &  & \ding{55} & \textcolor{blue!90!black}{2492$\times$20 QA} & SAQA & -\\
            \cmidrule(lr){1-7}
            UNCD~\citep{lang2025beyond} & \checkmark &  & \ding{55} & \textcolor{red!90!black}{2.9M+3.3M tokens} & 36K MCQA & - \\
            \bottomrule
        \end{tabular*}
        \label{tab:benchmarks}
        \subcaption{Benchmarks and datasets in unlearning.}
    \end{minipage}
    \begin{minipage}{.99\textwidth}
        \centering
        \begin{tabular*}{\linewidth}{@{\hspace{5pt}\extracolsep{\fill}} l l l m{100pt} @{\hspace{5pt}}}
            \toprule
            Benchmark & Field & Data & Used by\\
            \toprule
            {CounterFact}~\citep{meng2022locating} & Editing & 21,919 counterfactual records &  \citep{Wang2024Large,guo2024mechanistic,Patil2025UPCORE} \\
            \cmidrule(lr){1-4}
            {\scriptsize PKU-SafeRLHF}~\citep{ji2024pkusaferlhf} & Safety & QA pairs (265K w/ meta-labels, 166.8K w/ preference) & \citep{jia2024wagle,zhang2025resolving,Jia2024SOUL,Pan2025MultiObjective,chen2024wpn,chen2024machine,Liu2024Safer,Yao2024Large} \\
            \cmidrule(lr){1-4}
            SQuAD~\citep{rajpurkar2016squad} & Comprehension & 100K+ QA pairs of reading and reasoning & \citep{Pawelczyk2024Context,rezaei2024restor} \\
            \cmidrule(lr){1-4}
            ZsRE~\citep{levy2017zero} & {\scriptsize Relation extraction} & 30M+ positive examples, 2M+ negative examples & \citep{wang2025erasing,Wang2024Large} \\
            \bottomrule
        \end{tabular*}
        \subcaption{Benchmarks and datasets in relevant fields.}
        \label{tab:db_others}
    \end{minipage}
    \caption{Select a suitable benchmark. Part (a) organizes advantages of different tasks formats, data content and experiment paradigm. Part (b) outlines benchmarks and datasets with their data content (real or/and fictional (Fic.)), experiment paradigm (with or without fine-tuning (FT)), statistics of data (text/QA, `+' distinguishes between the unlearn set and the retain set), evaluation tasks (``Conti'': Continuation, ``SAQA'': Short answer QA, ``MCQA'': Multiple choice QA) and applications in subsequent studies. Part (c) outlines benchmarks in relevant fields with brief descriptions of their data.}
    \label{tab:D&B}
\end{table}

\subsubsection{Existing Benchmarks and Datasets}
A direct motivation for unlearning research comes from a number of works that aim to remove parts of the pretraining corpus~\citep{yao2024machine,Ashuach2024REVS,Wang2024Selective,cha2024robust,Stoehr2024Localizing,pochinkov2024dissecting,Tamirisa2024Robust,belrose2023leace}. 
Among these, the Pile dataset~\citep{gao2020pile}, which is commonly used in pretraining LLMs such as Pythia~\citep{biderman2023pythia}, is one of the most frequently adopted. 
\citet{lm_extraction_benchmark} further introduced the Training Data Extraction Challenge (TDEC), a subset of 20,000 examples of The Pile that has been employed as an unlearn set in several studies~\citep{jang2022knowledge,Tang2024Learn,Lee2024Protecting,barbulescu2024each,Gao2024Ethos,dong2024undial,Wang2024Selective,kassem2023preserving}. 
However, a major challenge is that the pretraining data for many state-of-the-art LLMs are not publicly available, and different models often use different corpora, significantly limiting the applicability of such datasets.

To address diverse research needs, numerous benchmarks and datasets have been developed, varying in motivation and application. 
Some focus on unlearning specific content, such as security information~\citep{Li2024WMDP,lang2025beyond}, copyrighted material~\citep{Eldan2023Whos,Shi2024MUSE}, or private data~\citep{Liu2024Learning,Maini2024TOFU}. 
Others emphasize knowledge connectivity or semantic diversity to enhance unlearning robustness~\citep{Qiu2024PISTOL,wu2024evaluating,Hong2024Intrinsic,yang2025faithun}. 
Additionally, works such as \citep{rezaei2024restor,veldanda2024llm} explore continuous learning–unlearning settings. 
Unlearning evaluations also frequently adapt benchmarks from related fields such as model editing and LLM safety. 
We summarize the characteristics of existing benchmarks in Table~\ref{tab:D&B}.

\subsection{Metrics}
\label{sec:eval_metric}
%\ref{goal}
After applying a unlearning method to a model on a selected dataset, we need several suitable metrics to evaluate effectiveness.
Recalling the goal of unlearning in Section~\ref{sec:def_goal}, the first kind of metric examines the \textbf{knowledge memorization} (Section \ref{sec:eval_mem}) of the unlearned model on the content of the unlearn set and the retain set.
Due to the various capabilities of LLMs, such as language proficiency and reasoning ability, the unlearned model should still retain all the \textbf{model utility} (Section \ref{sec:eval_util}).
Additionally, we expect the unlearning process is \textbf{robust} (Section \ref{sec:eval_adversary}) and \textbf{efficiency} (Section \ref{sec:eval_efficiency}).
%(1) One compares the unlearned model with the ideal unlearning result (expected model), that is, the evaluation of the result. 
%From different objects, these result-oriented metrics can be further divided into evaluating \textbf{knowledge memorization}, \textbf{model utility}, and \textbf{unlearning robustness}.
%(2) Another directly evaluates the unlearning process, mainly considering the \textbf{efficiency} of implementing the unlearning algorithm.
Refer to Figure \ref{fig:eval_frame} for an illustration of different metrics.

\subsubsection{Knowledge Memorization}
\label{sec:eval_mem}
In most cases, the ideal result of unlearning is expected to contain all of the retain set and none of the unlearn set.
The first kind of metrics evaluates \textbf{knowledge memorization}, which examines whether certain data have been memorized in the model.
Typically, the choice of knowledge memorization metric is tailored to the task format; for instance, accuracy is a direct measure for multiple choice questions.
In general, metrics are categorized into three distinct classes according to their operational basis: those applied to the model's final outputs, those applied to the model's internal logits, and membership inference attack (MIA).
We summarize the knowledge memorization metrics introduced in this section in Table \ref{tab:sim_metric}.

% Evaluation metrics are essential to quantify model performance on specific tasks and to assess the efficacy of machine unlearning algorithms. 
% Typically, the choice of metric is tailored to the task format; for instance, accuracy is a direct measure for multiple choice questions. 
% In general, metrics are categorized into two distinct classes according to their operational basis: those applied to the model's final outputs (similarity-based) and those applied to the model's internal logits (probability-based).
% We illustrate all the metrics introduced in this section in Figure \ref{fig:metric}.

%\textbf{Output-based.}
\paragraph{Output-based}
The most direct approach is have the model complete the selected task and compare the output with the ground truth.
Given that a model's output may not perfectly align with ground-truth references, multiple metrics are employed to quantify the textual similarity between them. 
%Commonly adopted metrics include verbatim matching, BLEU~\citep{papineni2002bleu}, ROUGE~\citep{lin2004rouge}, cosine similarity, and inference-based evaluation using Natural Language Inference (NLI) models. 
%The application of these different similarity metrics across the literature is summarized in Table~\ref{tab:sim_metric}.
\textbf{Verbatim matching} represents the simplest and most computationally efficient approach, particularly suitable for short or categorical answers. 
For longer and more complex generations, studies such as \citep{thaker2024guardrail,Wang2024Large} relax the exact match criterion to require the strict inclusion of specific keywords in the outputs. 
This adaptation demonstrates strong performance on benchmarks like TOFU~\citep{Maini2024TOFU}, where questions are often centered on a unique, identifiable entity (e.g., an author's name).
%Another way to relax is to check if there is a target token among the top-k tokens in the next token prediction process, and calculate metrics such as the top hit ratio (THR)~\citep{Qiu2024PISTOL}.
Another way to relax is to prompt the first $n$ tokens and check the $n+1$ token at each time, including memorization accuracy (MA)~\citep{tirumala2022memorization} and extraction strength (ES)~\citep{carlini2021extracting}.
\textbf{BLEU}~\citep{papineni2002bleu} and \textbf{ROUGE} (primarily ROUGE-N and ROUGE-L)~\citep{lin2004rouge} are established NLP metrics that focus on precision and recall of n-gram or longest common subsequence (LCS), respectively.
Beyond a single score, Extraction Likelihood (EL)~\citep{jang2022knowledge} measures extraction risk as the average ROUGE score across varying prefix lengths.
%To measure extraction risk, extraction likelihood (EL)~\citep{jang2022knowledge} prompt model with varying length of prefix and take an average ROUGE between output and suffix.
However, both ROUGE and EL treat all words with equal weight. 
To prioritize key information, \citet{xu2025relearn} introduced the Entity Coverage Score (ECS), which extracts key entities by an LLM and calculates similarity based solely on these entities.
%\textcolor{red}{Others: edit distance~\citep{russinovich2025obliviate}.}

In addition to the model-free metrics, some methods introduce external models for a better evaluation beyond lexical overlap.
Among these model-based methods, \textbf{BertScore}~\citep{zhang2019bertscore} constitutes a major category, typically involving the conversion of text into embedding vectors and then the calculation of cosine similarity. 
Through this embedding transformation, the model can better handle semantic-level information, such as synonyms, negation words, and word order.
For scenarios requiring more knowledge and understanding (such as recognizing that "born in London" and "born in the UK" are consistent), evaluation using \textbf{customized models}, such as Natural Language Inference (NLI) models~\citep{bos2005recognising} or BLEURT~\citep{sellam2020bleurt}, can yield more accurate results. 
Furthermore, more universal evaluation methods include \textbf{human evaluation} or \textbf{external LLM assessments}.
These methods are not only adaptable across diverse tasks, but can also comprehensively evaluate the wording and grammar of the outputs. 
However, they often function as black boxes and can be susceptible to biases inherent in human evaluators or proxy LLMs. 
Finally, several less prevalent metrics have also been applied in unlearning evaluations~\citep{Liu2024Large,Gu2024MEOW}, including METEOR~\citep{banerjee2005meteor}, MAUVE~\citep{pillutla2021mauve}, and Rep$_3$~\citep{welleck2019neural}.

%\textbf{Logit-based.}
\paragraph{Logit-based}
The autoregressive nature of Large Language Models (LLMs) involves computing a probability distribution for the next token conditioned on the preceding sequence. 
This \textbf{next token probability} can thus serve as an indicator of the model's latent knowledge for a given prompt, where a higher probability signifies stronger retention of that specific token~\citep{Eldan2023Whos,Maini2024TOFU}. 
A representative indicator calculated on this basis is perplexity, which is adopted as a metric in several studies~\citep{yao2024machine,doshi2024does,Ji2024Reversing,Wu2023DEPN,choi2024cross}, under the premise that more firmly memorized knowledge typically yields lower perplexity.
Several work has calculated comprehensive indicators based on the next token probability~\citep{wang2025effective,Mekala2024Alternate,Lee2024Protecting}.
A significant advancement is the \textbf{Truth Ratio} introduced by \citet{Maini2024TOFU}, which quantifies the likelihood of a correct answer relative to incorrect alternatives for a given question. 
Theoretically, a model lacking specific knowledge should exhibit a negligible difference in probability between correct and incorrect answers. 
The efficacy of unlearning can be further statistically validated by applying tests like the KS-Test to compare the distribution of Truth Ratios between the unlearned model and an expected baseline. 
The truth ratio can effectively detect under- and over-unlearning at the distribution level.
%Due to its effectiveness, the Truth Ratio has become a widely adopted metric in LLM unlearning evaluations~\citep{zhang2025resolving,wang2025erasing,cha2024robust,thaker2024guardrail,Huang2024Offset,Liu2024Large,Ji2024Reversing,jia2024wagle,Gu2024MEOW,choi2025opt-out}.

A known limitation of direct probability usage is the extreme variance in conditional probabilities across tokens, which can adversely affect metric stability. 
A straightforward mitigation is to utilize token \textbf{ranks} instead of raw probabilities. 
Sorting tokens by their probability in descending order and using the rank as the score results offer a more uniform score distribution~\citep{Ashuach2024REVS}. 
This rank-based paradigm is also employed by several metrics adapted for unlearning evaluation~\citep{Qiu2024PISTOL,choi2024cross}. 
For instance, Exposure~\citep{carlini2019secret}, a key metric in memorization analysis, can be viewed as a rank-based variant of the Truth Ratio, substituting likelihood comparison with rank comparison. 
Similarly, the Mean Reciprocal Rank (MRR), prevalent in entity retrieval tasks~\citep{lacroix2018canonical}, calculates the reciprocal average of the ranks of target tokens.

%\textbf{Other approach.}
\paragraph{Membership inference attack (MIA)}
In addition to the metrics above, some articles conduct MIA for evaluation~\citep{Jin2024RWKU,Shi2024MUSE,ramakrishna2025lume,Jia2024SOUL}. 
MIA is a privacy attack that determines whether specific data samples are part of a model's training set~\citep{shokri2017membership}, including
%Several MIA methods are used to evaluate the quality of LLM unlearning ~\citep{Jin2024RWKU,Shi2024MUSE,ramakrishna2025lume,Jia2024SOUL}, including 
LOSS~\citep{yeom2018privacy}, Zlib Entropy~\citep{carlini2021extracting}, Min-K\% Prob~\citep{shi2023detecting} and Min-K\%++ Prob~\citep{zhang2024min}.
However, although widely used in traditional unlearning evaluations, MIA is considered unsuitable for the LLM context, as it typically requires training numerous shadow models, which is both data-prohibitive and computationally intractable for LLMs~\citep{Liu2024Large,chen2024machine,Yao2024Large}.

\begin{table}[t]
    \footnotesize
    \centering
%    \begin{tabular*}{\linewidth}{@{\hspace{5pt}\extracolsep{\fill}} l@{\hspace{2pt}}l@{\hspace{2pt}}l m{150pt} m{120pt} @{\hspace{5pt}}}
    \begin{tabular*}{.99\linewidth}{@{\hspace{5pt}\extracolsep{\fill}} l l l m{140pt} m{120pt} @{\hspace{5pt}}}
        \toprule
        Obj. & Class & Name & Note (Advantages) & Used by\\
        \toprule
        \multirow{7}[30]{*}{\makecell[l]{Output}} & \multirow{3}[16]{*}{\makecell[l]{Model-\\free}} & Verbatim & Simple, computationally efficient, strong performance on some benchmarks.\newline\textcolor{red!90!black}{$\rightarrow$ keyword~\citep{thaker2024guardrail,Wang2024Large}, ES~\citep{carlini2021extracting}, MA~\citep{tirumala2022memorization}} & \citep{ramakrishna2025lume,cha2024robust,thaker2024guardrail,takashiro2025answer,Stoehr2024Localizing,pochinkov2024dissecting,Wang2024Large,Tian2024Forget,wang2025effective,Wang2024Rethinking,jang2022knowledge,Tang2024Learn,Lee2024Protecting,choi2024cross,barbulescu2024each,dong2024undial,Wang2024Selective} \\
        \cmidrule(lr){3-5}
        & & BLEU & Commonly used in translation (precision). \newline\textcolor{red!90!black}{$\rightarrow$ Scare-BLEU~\citep{post2018call}} & \citep{baluta2024unlearning,wu2024codeunlearn,hu2024separate,Liu2024Large,Ji2024Reversing,Gu2024MEOW,Scholten2024Probabilistic,Wang2024LLM,Tang2024Learn,Yao2024Large,Wang2023KGA} \\
        \cmidrule(lr){3-5}
        & & ROUGE & Commonly used in text summary (recall). Include LCS, ROUGE-L and ROUGE-N. \newline\textcolor{red!90!black}{$\rightarrow$ EL~\citep{jang2022knowledge}, ECS~\citep{xu2025relearn}} & \citep{Maini2024TOFU,ramakrishna2025lume,Qiu2024PISTOL,joshi2024towards,jiang2025holistic,yuan2024closer,wei2024underestimated,Scholten2024Probabilistic,wang2025erasing,shen2025lunar,zhang2025resolving,cha2024robust,hu2024separate,Chen2023Unlearn,ren2025general,Huang2024Offset,Liu2024Large,Ji2024Reversing,sanyal2025agents,wang2024when,Dou2024Avoiding,jia2024wagle,xu2025relearn,Gu2024MEOW,choi2025opt-out,ma2025unveiling,zhang2025catastrophic,Yuan2024Robust,wang2025effective,Mekala2024Alternate,Patil2025UPCORE,russinovich2025obliviate,Wei2024Evaluating,Liu2024Revisiting,Bu2024Unlearning,Sinha2024UnStar,Wang2024LLM,fan2025simplicity,jang2022knowledge,Tang2024Learn,Lee2024Protecting,Wang2024RKLD,Zhang2024Negative,Gao2024Ethos,dong2024undial,Wang2024Selective,zhang2025rule} \\
        \cmidrule(lr){2-5}
        %BertScore & - & - \\
        & \multirow{4}[12]{*}{\makecell[l]{Model-\\based}} & \makecell[l]{BertScore~\citep{zhang2019bertscore}} & Semantic information (synonym, etc.). \newline\textcolor{red!90!black}{$\rightarrow$ Calculated with other encoders} & \citep{yuan2024closer,wei2024underestimated,wu2024codeunlearn,Liu2024Large,sanyal2025agents,xu2025relearn,russinovich2025obliviate,Wei2024Evaluating,Tang2024Learn,Yao2024Large} \\
        \cmidrule(lr){3-5}
        & & \makecell[l]{Customized\\model eval} & Knowledge and understanding.\newline\textcolor{blue}{e.g. NLI, BLEURT~\citep{sellam2020bleurt}} & \citep{Liu2024Learning,jiang2025holistic,yuan2024closer,xu2025relearn,chen2024machine,Liu2024Safer,Yao2024Large} \\
        \cmidrule(lr){3-5}
        & & Human eval & Adaptable across diverse tasks, comprehensive in multiple dimensions. & \citep{liu2021dexperts,xu2025relearn,Tang2024Learn,Lu2022QUARK} \\
        \cmidrule(lr){3-5}
        & & LLM eval & Similar to the notes of human eval. & \citep{ramakrishna2025lume,lang2025beyond,hu2024separate,sanyal2025agents,wang2024when,xu2025relearn,Mekala2024Alternate,Liu2024Revisiting,Pan2025MultiObjective,Sinha2024UnStar,Tang2024Learn,Liu2024Safer,Lu2022QUARK,zhang2025rule}\\
        \midrule
        \multirow{3}[8]{*}{\makecell[l]{Logit}} & \multirow{2}[4]{*}{Prob.} & \makecell[l]{Next token\\probability} & Examination of the model's latent knowledge. \newline\textcolor{blue}{e.g. $p(a|q)$~\citep{Maini2024TOFU}, perplexity} \newline\textcolor{red}{$\rightarrow$ KL divergence~\citep{wang2025effective}, CI~\citep{Mekala2024Alternate}, RMA~\citep{Lee2024Protecting}} & \citep{Eldan2023Whos,Maini2024TOFU,yao2024machine,doshi2024does,Ji2024Reversing,Wu2023DEPN,wang2025effective,Mekala2024Alternate,Patil2025UPCORE,fan2025simplicity,Lee2024Protecting,choi2024cross,Wang2024RKLD,Zhang2024Negative,Wang2023KGA,Lu2022QUARK} \\
        \cmidrule(lr){3-5}
        & & Truth ratio~\citep{Maini2024TOFU} & Detect under- and over-unlearning at the distribution level. & \citep{zhang2025resolving,wang2025erasing,cha2024robust,thaker2024guardrail,Huang2024Offset,Liu2024Large,Ji2024Reversing,jia2024wagle,Gu2024MEOW,choi2025opt-out,Mekala2024Alternate,Jia2024SOUL,Wang2024Rethinking,Patil2025UPCORE,Bu2024Unlearning,Wang2024LLM,fan2025simplicity,Wang2024RKLD,Zhang2024Negative}\\
        \cmidrule(lr){2-5}
        & Rank & Rank score & Uniform score distribution, easy for comparison. \newline\textcolor{blue!90!black}{e.g. Exposure~\citep{carlini2019secret}, MRR~\citep{lacroix2018canonical}, THR~\citep{Qiu2024PISTOL}, PA~\citep{choi2024cross}} & \citep{Ashuach2024REVS,Wu2023DEPN,baluta2024unlearning,Qiu2024PISTOL,choi2024cross,Gu2024SecondOrder}\\
        \bottomrule
    \end{tabular*}
    \caption{Statistics of the use of different knowledge memorization metrics in LLM unlearning evaluations. \textcolor{red!90!black}{Red text} marks the new method improved on the corresponding method. \textcolor{blue!90!black}{Blue text} marks representative examples from the corresponding methods.}
    \label{tab:sim_metric}
\end{table}

\subsubsection{Model Utility}
\label{sec:eval_util}
Beyond investigating model memorization, various methodologies are employed to assess the general utility of unlearned models.
Some directly and efficiently computable indicators quantify specific aspects of model performance, which is referred to as \textbf{utility metrics}.
Among these, perplexity is one of the most frequently used measures; a lower perplexity signifies better fluency~\citep{liu2021dexperts}, higher model confidence~\citep{wei2024underestimated}, and improved meaningfulness of the generated content~\citep{Liu2024Large}.
In addition to perplexity, numerous studies focus on lexical diversity, proposing metrics such as the mean number or proportion of distinct n-grams~\citep{liu2021dexperts,Sinha2024UnStar,Liu2024Revisiting}, the unique token ratio~\citep{Liu2024Large,Yao2024Large}, and token entropy~\citep{yuan2024closer,Tang2024Learn}.
As noted by \citet{yuan2024closer}, reduced vocabulary diversity often correlates with token repetition in model outputs, indicating poorer readability and weaker overall utility.
Also, established linguistic indices, including Brunet’s Index~\citep{brunet1978vocabulaire} and Honore’s Statistic~\citep{honore1979some}, have been applied to assess lexical richness in unlearning contexts~\citep{xu2025relearn}.

Meanwhile, comprehensive \textbf{general benchmarks} are commonly utilized to evaluate the overall performance of unlearning methods~\citep{Jin2024RWKU, Li2024WMDP}.
Table \ref{tab:banchmark} summarizes frequently adopted benchmarks and their usage statistics across different unlearning studies.
Integrated evaluation frameworks and toolkits, such as Language Model Evaluation Harness~\citep{gao2021framework}, facilitate the systematic application of these benchmarks for LLM unlearning evaluation~\citep{jia2024wagle, Ashuach2024REVS, Wang2024Large, Tian2024Forget,Kadhe2024Split}.

\begin{table}[t]
    \footnotesize
    \centering
    \begin{tabular*}{.99\linewidth}{@{\hspace{10pt}\extracolsep{\fill}} l l m{240pt} @{\hspace{10pt}}}
        \toprule
        Class & Name & Used by\\
        \toprule
        Reasoning & ARC~\citep{clark2018think} & \citep{Eldan2023Whos,yao2024machine,Liu2024Learning,yuan2024closer,Chowdhury2024Scalable,thaker2024guardrail,Huang2024Offset,Ji2024Reversing,wang2024when,Wang2024Large,Tian2024Forget,Gu2024MEOW,veldanda2024llm,choi2025opt-out,chen2024wpn,Tang2024Learn,jang2022knowledge,Lee2024Protecting,Wang2024RKLD,barbulescu2024each,Lu2024Eraser,Gu2024SecondOrder,dong2024undial,Wang2024Selective,kassem2023preserving}\\
        \cmidrule(lr){1-3}
        \multirow{2}[1]{*}{Math} & GSM8K~\citep{cobbe2021training} & \citep{yao2024machine,Ashuach2024REVS,yuan2024closer,hu2024separate,Wang2024Large,Lizzo2024UNLEARN} \\
        \cmidrule(lr){2-3}
        & MathQA~\citep{amini2019mathqa} & \citep{Gu2024SecondOrder,Tang2024Learn,jang2022knowledge,Lee2024Protecting,Wang2024RKLD,barbulescu2024each,chen2024wpn,Wang2024Selective,kassem2023preserving} \\
        \cmidrule(lr){1-3}
        \multirow{3}[3]{*}{Commonsense} & HellaSwag~\citep{zellers2019hellaswag} & \citep{Eldan2023Whos,Chowdhury2024Scalable,thaker2024guardrail,Huang2024Offset,Ji2024Reversing,Wang2024Large,choi2025opt-out,russinovich2025obliviate,chen2024wpn,Tang2024Learn,jang2022knowledge,Lee2024Protecting,Wang2024RKLD,barbulescu2024each,Lu2024Eraser,dong2024undial,Wang2024Selective,kassem2023preserving} \\
        \cmidrule(lr){2-3}
        & PIQA~\citep{bisk2020piqa} & \citep{Eldan2023Whos,Liu2024Learning,Chowdhury2024Scalable,Ji2024Reversing,Gu2024MEOW,choi2025opt-out,chen2024wpn,Tang2024Learn,jang2022knowledge,Lee2024Protecting,Wang2024RKLD,barbulescu2024each,Gu2024SecondOrder,dong2024undial,Wang2024Selective,kassem2023preserving} \\
        \cmidrule(lr){2-3}
        & WinoGrande~\citep{sakaguchi2021winogrande} & \citep{Eldan2023Whos,Liu2024Learning,Chowdhury2024Scalable,Huang2024Offset,choi2025opt-out,russinovich2025obliviate,chen2024wpn,Tang2024Learn,jang2022knowledge,Lee2024Protecting,Wang2024RKLD,barbulescu2024each,dong2024undial,Wang2024Selective,kassem2023preserving} \\
        \cmidrule(lr){1-3}
        \multirow{2}[2]{*}{Comprehension} & Lambada~\citep{paperno2016lambada} & \citep{Liu2024Learning,choi2025opt-out,chen2024wpn,Gu2024SecondOrder,Tang2024Learn,jang2022knowledge,Lee2024Protecting,barbulescu2024each,kassem2023preserving} \\
        \cmidrule(lr){2-3}
        & OpenBookQA~\citep{mihaylov2018suit} & \citep{Eldan2023Whos,gao2024large,Chowdhury2024Scalable,thaker2024guardrail,Huang2024Offset,Ji2024Reversing,choi2025opt-out} \\
        \cmidrule(lr){1-3}
        \makecell[l]{Universal\\ knowledge} & MMLU~\citep{hendrycks2020measuring} & \citep{Jin2024RWKU,Li2024WMDP,yao2024machine,Ashuach2024REVS,ramakrishna2025lume,lang2025beyond,che2025model,doshi2024does,yuan2024closer,farrell2024applying,hu2024separate,Chowdhury2024Scalable,ren2025general,thaker2024guardrail,Liu2024Large,sanyal2025agents,wang2024when,Dou2024Avoiding,hu2025falcon,guo2024mechanistic,huu2024effects,Tian2024Forget,choi2025opt-out,zhang2025catastrophic,lucki2024adversarial,Tamirisa2024Robust,russinovich2025obliviate,Wei2024Evaluating,Lizzo2024UNLEARN,Kadhe2024Split,deeb2025unlearning} \\
        \cmidrule(lr){1-3}
        Multi-turn & MT-Bench~\citep{zheng2023judging} & \citep{Li2024WMDP,lang2025beyond,che2025model,doshi2024does,thaker2024guardrail,Dou2024Avoiding,russinovich2025obliviate,Wei2024Evaluating} \\
        \cmidrule(lr){1-3}
        \makecell[l]{Mimic human\\ Falsehoods} & TruthfulQA~\citep{lin2022truthfulqa} & \citep{Jin2024RWKU,yuan2024closer,hu2024separate,Chowdhury2024Scalable,Tian2024Forget,zhang2025catastrophic,Yuan2024Robust,russinovich2025obliviate,chen2024machine,Liu2024Safer,Yao2024Large} \\
        \bottomrule
    \end{tabular*}
    \caption{Overview of general benchmarks widely used in unlearning evaluation. We broadly divide these benchmarks into seven classes and count the use of each benchmark.} %Refer to Table \ref{tab:banchmark_total} for a complete statistics.}
    \label{tab:banchmark}
\end{table}

\subsubsection{Unlearning Robustness}
\label{sec:eval_adversary}
Empirical studies indicate that many machine methods merely suppress the surface-level expression of specific knowledge, leaving the underlying representations vulnerable to various adversarial attacks~\citep{Lynch2024Eight,Hong2024Intrinsic,lucki2024adversarial}. 
To systematically assess robustness, a range of adversarial techniques from the security domain have been integrated into the evaluation of LLM unlearning~\citep{Jin2024RWKU,Shi2024MUSE,ramakrishna2025lume,Jia2024SOUL}, which we collectively refer to as attack techniques.
Commonly adopted methods include the following.
(1) \textbf{Attack on the input}, such as crafted jailbreak prompts~\citep{shen2024do}, AutoPrompt~\citep{shin2020autoprompt}, Greedy Coordinate Gradient (GCG)~\citep{zou2023universal}, Prompt Automatic Iterative Refinement (PAIR)~\citep{chao2023jailbreaking}.
%(1) \textbf{Attack on the input}, such as crafted jailbreak prompts~\citep{shen2024anything}, in-context learning adversarial attacks~\citep{wei2023jailbreak}, GCG~\citep{zou2023universal}, AutoPrompt~\citep{shin2020autoprompt}, BEAST~\citep{sadasivan2024fast}, PAIR~\citep{chao2023jailbreaking}, persona modulation~\citep{shah2023scalable}, JailbreakHub~\citep{shen2024do}, many-shot jailbreaking~\citep{anil2025manyshot}.
(2) \textbf{Attack on the hidden layer}, such as probing techniques~\citep{belinkov2022probing,alain2016understanding}, Logit Lens~\citep{nostalgebraist2020interpreting}, soft-prompt-based threats~\citep{schwinn2024soft,che2025model}, AnonAct~\citep{seyitouglu2024extracting}.
Further heuristic attacks have also been proposed for specific unlearning scenarios~\citep{Ashuach2024REVS,shen2025lunar,Du2024Textual}. 
%In addition, two specialized strategies tailored to the unlearning context are frequently employed: relearning~\citep{Lynch2024Eight,joshi2024towards} and Membership Inference Attacks (MIA)~\citep{shokri2017membership}. 
%Further heuristic attacks have also been proposed for specific unlearning scenarios. 
%A comprehensive overview of adversarial techniques and robustness enhancement strategies will be provided in Section~\ref{sec:dis_adver}.

In response to the characteristics of LLM unlearning tasks, a relatively unique robustness evaluation method, relearning~\citep{Lynch2024Eight,joshi2024towards}, is also frequently used.
\textbf{Relearning} evaluates an unlearned model by exposing it to a limited subset of the unlearned data. 
In in-context relearning, knowledge related to the unlearn set, such as book summaries or relevant background information, are included in the prompts when evaluating the unlearned LLM.
When relearning by fine-tuning, the model is full-parameter or LoRA fine-tuned on a small portion of the unlearn set or a related set.
Empirical studies consistently demonstrate that relearning can substantially degrade unlearning quality, causing the model to rapidly recapitulate a significant portion of unlearned knowledge from sparse cues~\citep{Lynch2024Eight,lermen2023lora, zhan2023removing, guo2024mechanistic,joshi2024towards,fan2025simplicity}, or begin to systematically avoid generating content related to the unlearning target, even when contextually prompted~\citep{joshi2024towards}.
More critically, fine-tuning on data with content or distribution similar to the unlearn set can also reverse the unlearning effects, restoring model performance to a level comparable to its state before unlearning~\citep{doshi2024does,deeb2025unlearning}.

% Some works have found that many unlearning methods only suppress the expression of specific knowledge from the surface, and various attack methods can easily recover the unlearned knowledge~\citep{Lynch2024Eight,Hong2024Intrinsic,lucki2024adversarial}.
% To improve robustness, a large number of adversarial techniques have been introduced into the LLM unlearning evaluation.
% Most of the work adopt widely used attack methods in the security field~\citep{Lynch2024Eight,Jin2024RWKU,Hong2024Intrinsic,Li2024WMDP,che2025model,sanyal2025agents}, such as black-box methods including crafted jailbreak prompts~\citep{shen2024anything}, in-context learning adversarial attack~\citep{wei2023jailbreak}, GCG adversarial attack~\citep{zou2023universal}, and white-box methods including probe~\citep{belinkov2022probing,alain2016understanding} and soft prompt threats~\citep{schwinn2024soft,che2025model}. 
% In response to the characteristics of LLM unlearning tasks, two relatively unique attack methods, relearning~\citep{Lynch2024Eight,joshi2024towards} and Membership Inference Attack (MIA)~\citep{shokri2017membership}, are also frequently used.
% Additionally, there are some heuristic attack methods designed for specific unlearning scenarios or methods, such as Logit-Lens Attack (LLA), Delta Attack (DA), Perturbation Attack (PA)~\citep{Ashuach2024REVS} and Layer Skip Attack, Reverse Direction Attack~\citep{shen2025lunar}.
% A detailed introduction of adversarial techniques and robustness enhancement will be in Section~\ref{sec:dis_adver}.

\subsubsection{Unlearning efficiency}
\label{sec:eval_efficiency}
While the majority of existing studies concentrate on the efficacy of unlearning, the resource overhead of deploying such algorithms under real-world constraints remains a crucial consideration, including both memory occupation~\citep{BannihattiKumar2023Privacy,Jia2024SOUL} and computational time consumption.
For time cost, a straightforward approach is to directly measure the algorithm's runtime during experiments~\citep{Chen2023Unlearn,Chowdhury2024Scalable,BannihattiKumar2023Privacy,Jia2024SOUL}. 
Since computational speed is closely tied to GPU performance, some studies convert raw runtime into GPU hours (i.e., number of GPUs × training hours) to facilitate comparison~\citep{ren2025general,Ji2024Reversing}.  
Nonetheless, fair cross-study comparisons remain challenging due to variations in experimental environments.  
The most reliable method is to execute all algorithms under controlled conditions, though this is often resource-intensive.  
Alternatively, several works estimate time consumption theoretically, using metrics such as floating-point operations~\citep{cha2024robust} or gradient computation budgets~\citep{wei2024underestimated}.  
However, discrepancies between theoretical estimates and actual runtime may arise due to differences in implementation and hardware optimization.

\section{Challenges and Future Directions}
\label{sec:dis}

\subsection{Challenges}
%The existing unlearning work of LLMs faces many challenges. 
% From a goal perspective, unlearning lacks a clear and universally accepted formal definition (Section~\ref{sec:dis_def}), leading to a lack of accurate evaluation (Section~\ref{sec:dis_eval}). 
% From a content perspective, unlearning methods are influenced by different languages (Section~\ref{sec:dis_lang}) and data (Section~\ref{sec:dis_data}). 
% From a practical deployment perspective, unlearning methods need to handle problems of different scales (Section~\ref{sec:dis_scal}) and sequential unlearning requests (Section~\ref{sec:dis_conti}).

\subsubsection{Definition and Evaluation of Unlearning}
\label{sec:dis_def}
In Section~\ref{sec:def}, we characterize the goal of LLM unlearning as ensuring that ``the unlearned model should no longer memorize information from the unlearn set while preserving all other knowledge.''
%However, two key issues are ambiguous, resulting in differences in the definition of unlearning across different articles, which further leads to a lack of accurate evaluation.
However, two key issues remain ambiguous, leading to divergent definitions of unlearning across the literature and, consequently, to inconsistent and imprecise evaluation practices.

\textbf{(1) How should memorization be defined and detected?}
Most studies assess memorization based on model output in specific tasks, yet disagree on the criteria for judging these outputs.
For the content related to the unlearn set, some works argue that the model should simply avoid generating such content~\citep{Eldan2023Whos}, while others require it to explicitly respond with ``I don’t know''~\citep{shen2025lunar}.
Another line of research proposes that the unlearned model should produce outputs similar to those of a hypothetical retrained model, such as giving a specific incorrect answer~\citep{rezaei2024restor}.
When direct output is insufficient, adversarial methods are sometimes employed to expose memorization.
However, such approaches face inherent limitations: Overly weak attacks may fail to detect memorization, whereas overly strong ones can force the model to generate arbitrary content, casting doubt on their reliability as auditing tools~\citep{chen2025soft}.
%Moreover, certain attack methods are considered ill-suited to the LLM context, such as MIA, which typically requires training numerous shadow models, thus being both data-prohibitive and computationally intractable for LLMs~\citep{Liu2024Large,chen2024machine,Yao2024Large}.
%For example, \citet{Liu2024Large} question the use of MIA in evaluating LLM unlearning, as MIAs typically require training numerous shadow models, which is a process that is both data-prohibitive and computationally intractable for LLMs.

\textbf{(2) What should constitute the unlearn set?}
For synthetic datasets such as TOFU~\citep{Maini2024TOFU}, this question is relatively straightforward.
However, in real-world scenarios, data interconnectivity complicate the identification of appropriate unlearning targets.
\citet{Tian2024Forget} adopt a legal perspective to determine which copyrighted or private data should be unlearned, while \citet{wu2024evaluating} construct relationship graphs to identify necessary data for removal.
Despite their merits, these methods remain reliant on manual, domain-specific analysis, and lack generalizability.

\subsubsection{Effect of Unlearning}
Unlearning has different effects on different languages and data, further increasing the difficulty of designing and evaluating unlearning algorithms.
%\subsubsection{Unlearning Effect across Languages}
%\label{sec:dis_lang}
%As pointed out by several works~\citep{yong2023low, deng2023multilingual}, inputs with low-resource languages are more likely to bypass the GPT-4's safeguard compared to the input of high-resource language, resulting in unsafe output. 
%Taking this as inspiration, 

\textbf{Effects across languages.}
Some studies conduct evaluations with prompts translated into languages other than English, finding that monolingual unlearning is fundamentally insufficient for multilingual LLMs~\citep{Lynch2024Eight,Jin2024RWKU}.
Furthermore, more languages that systematically divide into high- and low-resource are used in evaluations~\citep{choi2024cross,lu2025learn}, revealing the fact that unlearning in one language does not necessarily transfer to others and could even inadvertently reinforce harmful content across languages.
Together, these findings underscore a critical consensus: effective and secure unlearning necessitates multilingual joint unlearning strategies to address model behavior holistically in all languages.

%\subsubsection{Unlearning Effect across Data}
%\label{sec:dis_data}
%Data have multiple characteristics, such as data distribution and intrinsic correlation, which will affect the effect of unlearning.
\textbf{Effects across data.}
From the perspective of data distribution, \citet{baluta2024unlearning} demonstrate that out-of-distribution (OOD) data require more gradient ascent but offer a better unlearning quality, whereas in-distribution data allow faster unlearning but severely compromise model utility, illustrating a fundamental trade-off between unlearning efficiency and model preservation.
Considering the logical connectivity of the data, \citet{choi2024breaking} identify that current unlearning methods struggle with multi-hop knowledge, where unlearning one intermediate fact in a chain often fails to remove the entire logical sequence. 
%To address this, they propose MUNCH, a method specifically designed to unravel the links in multi-hop knowledge unlearning. 
Furthermore, some studies investigate the impact on adjacent data after performing unlearning on selected data, identifying phenomena called ``transfer unlearning''~\citep{lu2024transfer}, ``ripple effect''~\citep{zhang2025theft} and ``onion effect''~\citep{borkar2023learn}.
These effects highlight the intricate and unpredictable consequences of unlearning, emphasizing the need for careful monitoring to ensure that unlearning achieves its intended goals without introducing new risks.

\subsubsection{Unlearning in Reality}
%\subsubsection{Scalability}
%\label{sec:dis_scal}
A significant challenge lies in the \textbf{scaling gap} between experimental settings and real-world conditions. 
Current unlearning experiments are largely limited to models with fewer than 10 billion parameters and unlearning sets under 1 billion instances, raising concerns about the applicability of these methods to larger models and datasets.  
\citet{Shi2024MUSE} analyze how evaluation metrics evolve as the size of the unlearn set increases, providing insight into scalability.  
On the other hand, in practical deployments, large models are often compressed, such as using quantization for efficiency.  
Notably, \citet{zhang2025catastrophic} demonstrate that quantizing unlearned models can inadvertently reactivate unlearned knowledge, highlighting a key scalability challenge.

%\subsubsection{Continuous Unlearning}
%\label{sec:dis_conti}
In commercial applications, unlearning requests typically arrive sequentially, requiring models to \textbf{continuously unlearning} while maintaining performance~\citep{rezaei2024restor,veldanda2024llm}.  
To assess long-term viability, \citet{Shi2024MUSE} collect model checkpoints after processing each sequential request and track evaluation metrics over time.  
This approach helps quantify the cumulative impact of repeated unlearning and the model's ability to sustain utility.
Unfortunately, current unlearning methods are not yet ready to handle sequential unlearning.

\subsection{Future Directions}
\subsubsection{Unlearning in Specialized Architectures and Scenarios}
\label{sec:dis_arch}
The field is moving towards addressing unlearning in sophisticated model architectures. 
\citet{cheng2025tool} pioneer this effort for tool-augmented large language models (LLMs) by proposing ToolDelete, the first unlearning framework designed to remove a specific “skill” or the ability to use a particular tool, and they introduce a new membership inference attack (MIA) for evaluation. 
Similarly, the unique structure of Mixture-of-Experts (MoE) models presents a distinct challenge. 
\citet{zhuang2025seuf} find that unlearning a single expert is insufficient and propose the Selected Expert Unlearning Framework (SEUF) to effectively perform unlearning on MoE models. 
These works demonstrate that effective unlearning requires bespoke algorithms tailored to a model's specific architecture and knowledge organization.

\subsubsection{Unlearning as Tools}
Unlearning is not only a goal in itself, but also a powerful tool when we further expand the scope of unlearning targets.
Firstly, when choosing injected trojans or backdoor triggers as the target, unlearning can be an effective tool in defense~\citep{jiang2025backdoor,zhao2025unlearning,hernandez2024if,Kazemi2024Unlearning,Lu2024Eraser}.
Similarly, an opposite target, such as removing the safety alignment or disrupting the subsequent fine-tuning process on a base model, can convert unlearning to a means of attack~\citep{Rashid2024Forget}.
Furthermore, if unlearning the selected training data and examining the changes in the model before and after unlearning, we can have novel insights into how different data components contribute to and influence the final model capabilities~\citep{zhao2024deciphering,isonuma2024unlearning}.
A powerful and accurate unlearning method will play an important role as a tool.

%\citet{zhao2024deciphering} utilize GRACE, a data influence analysis method based on unlearning, to decipher the complex impact of various types of pretraining data on LLM performance. 
%This approach provides novel insights into how different data components contribute to and influence the final model capabilities.
%\citet{isonuma2024unlearning} propose UnTrac to trace training data that influence model outputs, helping to reduce the generation of harmful content.

\subsubsection{Unlearning beyond Data}
Most existing studies focus on unlearning specific data instances. 
However, in practical scenarios, unlearning requests often target not only concrete data but also abstract concepts or capabilities, such as erroneous reasoning patterns, harmful ethical values, or unsafe skills~\citep{lang2025beyond,Li2024WMDP}. 
Extending unlearning beyond the data to encompass abstract constructs is essential to prevent the propagation of incorrect or harmful knowledge. 
Achieving this goal may present two main pathways:
one is to precisely identify and modify parameters or representations associated with particular concepts or abilities; 
the other leverages established alignment techniques, such as reinforcement learning, by designing appropriate reward mechanisms that penalize the generation of undesirable content.

% Most research focuses on unlearning specific data.
% However, the targets of numerous unlearning requests are not specific data, but rather abstract concepts or abilities, such as wrong reasoning mode, harmful moral values or unsafe capabilities~\citep{lang2025beyond,Li2024WMDP}.
% Moving beyond unlearning specific data to abstract concepts or abilities is crucial to prevent the dissemination of wrong or dangerous knowledge.
% On the one hand, this may require precise modification by locating parameters or representations related to specific concepts or capabilities.
% On the other hand, by designing a suitable reward mechanism to penalize the model that generates undesirable content, this goal can be achieved through reinforcement learning or other mature alignment methods.

\subsubsection{Robust Unlearning}
In light of the observed fragility of LLM unlearning, a significant research direction aims to develop techniques that enhance its robustness and long-term stability. 
These defensive efforts pursue two primary objectives:
(1) to ensure that knowledge removal is thorough and persistent, thereby resisting attempts at recovery; and
(2) to prevent the unlearning procedure from introducing new vulnerabilities or unintended side effects.
Several existing studies address the first objective through robust unlearning frameworks~\citep{tamirisa2024tamper,zhang2024unforgettable,huu2025improving,fan2025towards} or methods that strengthen the robustness of unlearned models~\citep{zhang2024unforgettable,huu2025improving}. 
Nevertheless, given the proliferation of advanced attacks, achieving truly robust unlearning remains a critical and ongoing topic.

\subsubsection{Verifiable and Certifiable Unlearning}
In most current practices, unlearning is applied to models that have already internalized the content targeted for removal through opaque mechanisms, complicating the certification of unlearning effectiveness. 
However, from legal, safety, and social trust perspectives, achieving verifiable and trustworthy unlearning remains critically important.
To validate existing unlearning methods, it is essential to establish a fair and comprehensive evaluation benchmark. 
Looking ahead, future work may also draw inspiration from frameworks such as SISA by designing structured data storage and training protocols to enable intrinsically verifiable unlearning.

% In most cases, the existing LLM unlearning is performed on a model that already contains the content of the unlearn set through some inexplicable means, resulting in difficulties in the certification of unlearning effectiveness.
% However, from the perspectives of law, safety, and social trust, verifiable and trustworthy unlearning is relative critical. 
% For existing methods, a fair and comprehensive evaluation benchmark needs to be designed for certification.
% Meanwhile, future works can also draw on the ideas of SISA and design a reasonable data storage and training mechanism to ensure verifiable unlearning.
%\input{sec/6_applications}
%\input{sec/7_challenges}
% !TeX root = ../main_csur.tex
\section{Conclusions}
Machine unlearning has emerged as a pivotal technique to address critical challenges in large language models, including privacy protection, copyright compliance, and safety enhancement. 
%This approach enables the selective removal of specific data influences without the prohibitive cost of complete model retraining. 
In this survey, we provide a comprehensive review of work dedicated to LLM unlearning, including the definition and goal of LLM unlearning, the most recent LLM unlearning methods, and commonly used datasets and evaluation metrics of unlearning.
%Compared to existing surveys, 
Despite significant progress, the field of LLM unlearning remains in its early stages, with fundamental challenges
in the definition, evaluation, effects and practical deployment of unlearning.
Furthermore, we suggest several promising directions for future research.
We hope that this survey can provide readers with a general understanding of recent progress in this field and shed some light on future developments.
%\textcolor{red}{TBD here.}
%\input{sec/7_extra}

\bibliography{ref}
\bibliographystyle{ACM-Reference-Format}

\end{document}